\pdfoutput=1
\documentclass{article}

\usepackage[preprint]{neurips_2024}

\usepackage[utf8]{inputenc} 
\usepackage[T1]{fontenc}    
\usepackage{hyperref}       
\usepackage{url}            
\usepackage{booktabs}       
\usepackage{amsfonts}       
\usepackage{nicefrac}       
\usepackage{microtype}      
\usepackage{xcolor}         
\usepackage{amsmath}
\usepackage{amssymb}
\usepackage{nomencl}
\usepackage{glossaries}
\usepackage{xspace}
\usepackage{multirow}
\usepackage{graphicx}
\usepackage{bbding}
\usepackage{pifont}
\usepackage{algorithm}
\usepackage{subfigure}
\usepackage{enumitem}
\usepackage{multicol}
\usepackage{caption}
\usepackage{array}
\usepackage{fp}
\usepackage{makecell}
\usepackage{flushend}
\usepackage{cases}
\usepackage{color}
\usepackage{algpseudocode}
\usepackage{listings}
\usepackage{mathrsfs}
\usepackage{longtable}
\usepackage{colortbl}
\usepackage{bm}
\usepackage{tabularx}
\usepackage{caption}
\setlist[itemize]{noitemsep, topsep=0pt}
\definecolor{codegreen}{rgb}{0,0.3,0.6}
\definecolor{codegray}{rgb}{0.5,0.5,0.5}

\newcommand{\ie}{\emph{i.e.,}\xspace}
\newcommand{\eg}{\emph{e.g.,}\xspace}
\newcommand{\aka}{\emph{a.k.a.,}\xspace}

\newcommand{\paratitle}[1]{\vspace{1.5ex}\noindent\textbf{#1}}

\newcommand{\ignore}[1]{}

\usepackage{tcolorbox}
\definecolor{darkorange}{RGB}{255, 140, 0}
\definecolor{lightgreen}{RGB}{145, 204, 117}
\definecolor{lightyellow}{RGB}{250, 200, 88}
\definecolor{lightred}{RGB}{238, 102, 102}
\definecolor{lightblue}{RGB}{115, 192, 222}
\newtcolorbox{promptbox}[2][Prompt]{
colback=black!5!white,
arc=5pt, 
boxrule=0.5pt,
fonttitle=\bfseries,
title=#1, 
before upper={\scriptsize}, fontupper=\fontfamily{ptm}\selectfont,
colframe=#2, 
}

\title{An Empirical Study on Eliciting and Improving R1-like Reasoning Models}

\author{%
 Zhipeng Chen$^{1}$\thanks{Equal contribution.}~,
  Yingqian Min$^{1*}$,
  Beichen Zhang$^{1*}$,
  Jie Chen$^{1*}$,\\
  \textbf{Jinhao Jiang$^{1}$,~Daixuan Cheng,~Wayne Xin Zhao$^{1}$\thanks{Correspondence to Wayne Xin Zhao.}~, 
  Zheng Liu$^2$,} \\
  \textbf{Xu Miao$^3$,~Yang Lu$^3$,~Lei Fang$^3$,~Zhongyuan Wang}$^2$, \textbf{Ji-Rong Wen}$^{1}$
  \vspace{3pt} \\
  $^1$Gaoling School of Artificial Intelligence, Renmin University of China\\
  $^2$BAAI~~~~~
  $^3$DataCanvas Alaya NeW\\
  {\small\texttt{\{zhipeng\_chen,yingqianm,ptyzchenjie,jrwen\}@ruc.edu.cn}}\\
  {\small\texttt{\{zhangbeichen724,batmanfly\}@gmail.com}}
}

\begin{document}
\textit{Technical Report on Slow Thinking with LLMs: III}

\maketitle

\begin{abstract}
In this report, we present the third technical report on the development of slow-thinking models as part of the STILL project. As the technical pathway becomes clearer, scaling RL training has become a central technique for implementing such reasoning models. We systematically experiment with and document the effects of various factors influencing RL training, conducting experiments on both base models and fine-tuned models.  
Specifically, we demonstrate that our RL training approach consistently improves the \textsc{Qwen2.5-32B} base models, enhancing both response length and test accuracy. Furthermore, we show that even when a model like \textsc{DeepSeek-R1-Distill-Qwen-1.5B} has already achieved a high performance level, it can be further refined through RL training, reaching an accuracy of 39.33\% on AIME 2024.  
Beyond RL training, we also explore the use of tool manipulation, finding that it significantly boosts the reasoning performance of large reasoning models. This approach achieves a remarkable accuracy of 86.67\% with greedy search on AIME 2024, underscoring its effectiveness in enhancing model capabilities. 
We release our resources at the STILL project website:  \url{https://github.com/RUCAIBox/Slow_Thinking_with_LLMs}.

\end{abstract}

\section{Introduction}
\label{sec-intro}

Recently, large reasoning models (LRMs), \aka slow-thinking models, have achieved notable progress in advancing the slow, deliberate thinking capabilities of large language models (LLMs), enabling them to address complex tasks more effectively~\cite{zhao2023survey,deepseek-r1,kimik1.5}. 
By scaling test-time computation, these models generate extended thought processes before delivering a solution to a given problem. Unlike train-time scaling, which incur a one-time additional cost, test-time scaling imposes an additional cost per query. In essence, it learns to trade off more token outputs for improved performance. To achieve these performance improvements, the model must learn to effectively generate and employ critical reasoning steps, such as verification and reflection, within its output. This ensures that the additional intermediate outputs meaningfully guide the model toward the correct final solution. 
Since this process involves assessing and exploring alternative solutions within the output window, it effectively conducts a solution search within the natural language space. 
This reasoning paradigm represents a notable departure from earlier approaches in LLMs, such as short chain-of-thought~(CoT) reasoning, marking a substantial evolution in the field.

Since OpenAI initially released {o1-preview}, an early version of LRMs, the research community has shown immense enthusiasm for exploring the underlying technical pathways of these models. Despite numerous efforts to develop complex models~\cite{jiang2024technical,qiu-o1-survey}, the true direction of progress became evident with the release of the technical report and the open-sourcing of the reasoning model by {DeepSeek-R1}~\cite{deepseek-r1}. Specifically, the breakthrough was achieved by scaling reinforcement learning~(RL) training, which incentivizes LLMs to learn such thought processes through self-exploration. 
When the actual technical details are unveiled, it becomes fascinating to reflect on how RL has driven the advancement of LRMs throughout this progress. An early  notable example connecting RL with LLMs dates back to OpenAI's work on integrating human preferences to improve specific text generation tasks~\cite{learn-to-summarize}. Building on this, they employed a similar RL technique by utilizing human feedback to align LLMs with human preferences, culminating in the development of the groundbreaking work {InstructGPT}~\cite{Ouyang2022instruct}. This approach is formally termed reinforcement learning from human feedback (RLHF).  Despite the performance gains achieved through RLHF, a significant limitation is its reliance on high-quality human annotations to train an effective reward model, which acts as a proxy for human annotators by providing intermediate feedback during model training. However, training such a reward model has proven to be challenging, particularly when aiming for general-purpose applicability across all domains. 

So, how do LRMs overcome this challenge? The solution is surprisingly straightforward: leveraging verifiable problems (\eg math problems) that provide direct ground-truth answers. When a model generates a solution, the final answer is parsed and compared against the ground-truth one. We can simply assign pre-defined reward scores according to the comparison results. 
Another source of such training data is coding problems, where the correctness of the generated code can be automatically verified through unit tests. By utilizing these verifiable problems for RL training, we actually employ rule-based rewards. Actually, AlphaGo has also used rule-based rewards instead of training the reward models. 
While this idea is compelling, a notable limitation arises in domains where verifiable problems are absent. The question, then, is how this approach can be effectively applied to such domains. This challenge can be partially addressed through two key observations\footnote{An important and promising research direction is to investigate the development of more principled reinforcement learning (RL) training approaches that utilize general-domain rewards to tackle complex reasoning tasks. We leave this challenging topic for future work.}. First, it has been observed that long CoT reasoning embodies a universal thinking mode, which, once mastered in a specific domain, can be generalized across other domains. This phenomenon has been explored in our prior research on long CoT data distillation~\cite{still2}. Second, the reasoning capabilities of a model can be further enhanced through traditional post-training processes. For instance, as demonstrated by {DeepSeek-R1}, the joint utilization of rule-based and trained reward models enables effective all-domain alignment, thereby extending the applicability of the approach to domains lacking verifiable problems. 

When setting up the reward, it is crucial to discuss the RL algorithms used in developing LRMs. In practice, tuning and successfully running RL algorithms is challenging due to the training complexity, both in terms of the algorithmic systems (involving additional models) and optimization techniques (requiring more hyperparameters to tune). Therefore, even if the path is clear, directly replicating {DeepSeek-R1} is not straightforward, particularly when the training budget is limited. By reviewing recent studies on  scaling RL training~\cite{deepseekr1,kimik1.5}, three major features can be observed as training progresses: increasing training rewards, increasing response length, and emergent reasoning patterns. These factors are key indicators of the success of scaling RL training. To date, there have been numerous studies that attempt to replicate {o1} and {R1}, which provide important technical insights for developing LRMs. 
Our work largely builds on these prior efforts and aims to contribute independently to this critical exploration.

For this purpose, we initiated the exploration of scaling RL training for LRMs at the end of December 2024 as part of our slow-thinking project ``STILL''. We extensively experimented with various feasible and potential approaches to implementing reasoning models. This technical report aims to document the observations and insights gained from a large number of RL experiments. While we do not claim novelty or original contributions, our aim is to consolidate the lessons and insights gained throughout this process. We regard this exploration as a valuable learning experience, deepening our team's understanding of RL training for LLMs.
Concretely, in this report, we first thoroughly investigate the impact of RL settings on training effectiveness.
Next, we directly incentivize the base model through RL training to develop complex reasoning capabilities, observing that the model gradually spends more time ``thinking'' and exhibits advanced reasoning behaviors (\eg verification or reflection).
Finally, to further enhance the reasoning abilities of the fine-tuned model, we explore both RL and tool augmentation as strategies to improve the model's reasoning performance, achieving significant improvements across both small-sized (1.5B) and medium-sized LLMs (32B).

Based on our experiments, our major findings are summarized below:

$\bullet$ 
The performance of large reasoning models is heavily influenced by the settings of RL. We thoroughly investigate and document these effects by testing a range of parameter configurations  (Table~\ref{tab:recommandation}). Notably, the on-policy learning strategy proves to be a crucial element, enabling the model to achieve consistent performance improvements throughout the RL process and demonstrating its substantial impact on enhancing RL training (Section~\ref{sec:rl_setting}). 

$\bullet$  After pre-training, the base models already exhibit the potential to perform individual complex reasoning actions. The RL process effectively activates this capability, enabling the model to integrate these actions into a coherent and deliberate thinking process. Our RL training approach consistently improves the \textsc{Qwen2.5-32B} base model, enhancing both response length and test accuracy  (Section~\ref{sec-still_3_zero}). 


$\bullet$ Response length serves as an important indicator of the success of RL training; however, it is a consequence, not a cause, of performance improvement.  
Designing specialized reward functions to explicitly encourage the model to produce longer responses may lead to issues such as reward hacking, which can't inherently enhance the model's reasoning capabilities  (Section~\ref{sec-finetuned_longcot_data} and Section~\ref{sec:Auxiliary Approaches in RL}).

$\bullet$ RL training consistently improves the performance of fine-tuned models, encompassing both short and long CoT reasoning models. Even after Qwen2.5-1.5B attains a high level of performance through training with distilled data, RL training further elevates its capabilities, achieving a remarkable accuracy of 39.33 on AIME 2024 (Section~\ref{sec-still-3-1.5}).


$\bullet$ Through supervised fine-tuning, LRMs can acquire the capability to manipulate external tools, leading to a significant enhancement in the model's performance. By effectively utilizing tool manipulation, \textsc{Still-3-Tool-32B} achieves an impressive accuracy of 86.67 (greedy search) on AIME 2024. Remarkably, this ability can be activated with only a small number of high-quality training instances (Section~\ref{sec-still-3-tool}).

We release the necessary resources to reproduce our results at the STILL project website:  \url{https://github.com/RUCAIBox/Slow_Thinking_with_LLMs}.

\section{Experimental Settings}~\label{sec:exp_setting}
\label{sec-exp_setting}

In this section, we set up the experiments, and introduce the training framework, training data, backbone model, reward design and evaluation benchmarks.

\paratitle{Training Framework.}
We conduct experiments using two popular open-source repositories for reinforcement learning of language models: OpenRLHF~\cite{openrlhf} and veRL~\cite{verl}. For training \textsc{STILL-3-1.5B} in Section~\ref{sec-still-3-1.5}, we leverage OpenRLHF to implement the major code. For all other experiments, including those from the base model in Section~\ref{sec-zero-model} and from the fine-tuned models in Section~\ref{sec-code-start-model}, we employ veRL for developing our experiments.



\paratitle{Backbone Models.}
We experiment with various versions of \textsc{Qwen2.5} models~\cite{qwen2.5}. In our ``Zero'' experiments in Section~\ref{sec-zero-model}, we train from the base models of \textsc{Qwen2.5} series. For fine-tuned models in Section~\ref{sec-code-start-model}, we train from 1.5B and 32B \textsc{Qwen2.5} models of the \textsc{DeepSeek-R1-Distill} series~\cite{deepseekr1}. Additionally, we conduct experiments on fine-tuned models where the fine-tuning data are synthesized ourselves.

\paratitle{Training Data.}
We curate our training data according to the following principles: (1) \textit{Diversity:} We collect diverse sources, including AIME (up to 2023), MATH~\cite{dan2021math}, NuminaMath~\cite{li2024numinamath}, and Open Reasoner Zero~\cite{orz}; (2) \textit{Verifiability:} We remove examples of multiple-choice questions, proof questions, concept questions, open-ended questions, and questions with multiple sub-problems due to challenges in answer verification. We then filter out data examples where the answers cannot be parsed into digits using the SymPy library~\cite{aaron2017sympy}; (3) \textit{Difficulty:} We leverage a model-based filtering strategy that removes problems on which \textsc{Qwen-7B-Instruct}~\cite{qwen2.5} achieves either too high or zero pass rates. The resulting dataset contains 90k examples.





\paratitle{Reward Design.}
We design and validate a diverse set of rewards and analyze their effects on model performance, including output reward, format reward, length reward, and action reward. First, \emph{output reward} evaluates whether the final answer matches the ground truth. We set the reward to $1$ if the answer is correct and $0$ otherwise. Additionally, if the model fails to place its final answer in the \texttt{\textbackslash boxed\{\}}, the reward is set to $0$. Second, we employ \emph{format reward}~\cite{deepseekr1} in our ``Zero'' experiments in Section~\ref{sec-zero-model} to guide base models in structuring their responses correctly. Moreover, we explore novel auxiliary rewards in Section~\ref{sec:Auxiliary Approaches in RL}, including \emph{length reward}, which encourages longer responses, and \emph{action reward}, which incentivizes complex reasoning actions.

\paratitle{Evaluation Benchmarks.}
We evaluate model performance on various mathematical reasoning tasks, including MATH-OAI~\cite{dan2021math}, AIME, Omni-MATH~\cite{OmniMATH}, LiveAOPs~\cite{aopsdataset}, and HMMT. MATH-OAI consists of $500$ competition-level mathematics problems from the MATH~\cite{dan2021math} test set. AIME presents $30$ challenging problems per year for top high school students. We primarily evaluate on AIME 2024 throughout the experiments, and AIME 2025 is additionally evaluated for \textsc{STILL-3-Tool-32B}. Omni-MATH is an Olympiad-level benchmark comprising $4,428$ competition problems. LiveAOPs utilizes posts from the AoPS forum to create a contamination-resistant evaluation set of $3,863$ examples. HMMT is a high school mathematics competition, and we use the evaluation split from February 2025 provided by MathArena~\cite{matharena}, which contains 30 problems.

\paratitle{Compute Environment.}
Our experiments are primarily conducted on the Alaya NeW AI Operating System, DataCanvas' flagship computational orchestration platform. This platform enables cross-cluster heterogeneous resource unification, intelligent scheduling across distributed data centers, and seamless integration with a serverless framework featuring dynamic elastic scaling.


\begin{table}[t]
\small
    \centering
    \caption{Our exploration and suggestions of the settings of RL training. Please note that these recommendations are derived from our limited experiments and practical experiences, and they may not necessarily represent the optimal configurations. }
    \begin{tabular}{lcc}
        \toprule
        \textbf{Factor} & \textbf{Explored Options} & \textbf{Recommended Option}  \\
        \midrule
        \multirow{2}*{Train Batch Size} & $TBS=128$ & \multirow{2}*{\begin{tabularx}{0.45\textwidth}{@{}X@{}}\centering $TBS=1024$\\(Enhancing the training efficiency)\end{tabularx}}  \\
        & $TBS=1024$ &  \\
        \midrule
        \multirow{2}*{Learning Strategy} & On-policy Training & \multirow{2}*{\begin{tabularx}{0.45\textwidth}{@{}X@{}}\centering On-policy Training\\(Enhancing the training efficiency)\end{tabularx}} \\
        & Off-policy Training & \\
        \midrule
        \multirow{2}*{Rollout Times} & $n=8$ & \multirow{2}*{\begin{tabularx}{0.45\textwidth}{@{}X@{}}\centering $n=64$\\(Expanding exploration space)\end{tabularx}} \\
        & $n=64$ &  \\
        \midrule
        \multirow{3}*{Rollout Temperature} & $T=0.6$ & \multirow{3}*{\begin{tabularx}{0.45\textwidth}{@{}X@{}}\centering $T=1.2$\\(Expanding exploration space)\end{tabularx}} \\
        & $T=1.0$ & \\
        & $T=1.2$ & \\
        \midrule
        \multirow{3}*{Coefficient of KL} & $KL=0.001$ & \multirow{3}*{\begin{tabularx}{0.45\textwidth}{@{}X@{}}\centering Dynamic KL Annealing\\(Balancing constraints and exploration)\end{tabularx}} \\
        & $KL=0$ & \\
        & Dynamic KL Annealing & \\
        \midrule
        \multirow{4}*{Backbone Model} & \textsc{Qwen2.5-1.5B} & \multirow{4}*{\begin{tabularx}{0.45\textwidth}{@{}X@{}}\centering \textsc{Qwen2.5-32B}\\(Possessing stronger learning capacities)\end{tabularx}}  \\
        & \textsc{Qwen2.5-7B} & \\
        & \textsc{Qwen2.5-32B} & \\
        & \textsc{Qwen2.5-7B-Instruct} & \\
        \midrule
        \multirow{2}*{Training Prompt} & Simple Instruction & \multirow{2}*{\begin{tabularx}{0.45\textwidth}{@{}X@{}}\centering Detailed Instruction\\(Enhancing reasoning efficiency of the model)\end{tabularx}} \\
        & Detailed Instruction & \\
        \bottomrule
    \end{tabular}
    \label{tab:recommandation}
\end{table}

\section{RL Experiments on the Base Model}
\label{sec-zero-model}

Traditional post-training approaches often rely on SFT to initialize models before RL training, with the goal of establishing a robust foundation for effective RL optimization. 
Following DeepSeek-R1-Zero~\cite{deepseekr1}, in this section, we conduct experiments by directly applying RL to the pre-trained base models without any intermediate SFT stage.
This approach aims to explore whether LLMs can autonomously develop reasoning capabilities through pure RL-driven self-improvement. Our experiments systematically examine four critical dimensions. First, we empirically investigate the impact of training hyperparameters. Second, we analyze the effect of backbone models by comparing different base models and benchmarking them against fine-tuned models with short CoT reasoning capabilities. 
{Third, we delve into the impact of the prompt design on the reasoning ability of base models during RL training. 
}
{Last, we examine the emergence of representative reasoning patterns (\eg verification or reflection) during RL training. 
}
Furthermore, we adopt our findings on the RL training of \textsc{Qwen2.5-32B} and propose \textsc{STILL-3-Zero-32B}, which undergoes direct RL training without the SFT process.





\subsection{Exploring the Settings of RL Training}
\label{sec:rl_setting}
The effectiveness of RL training is highly influenced by a variety of factors. In this section, we concentrate on analyzing the impact of two critical aspects: hyper-parameter and training prompts. 
{
The following experiments are conducted based on the dataset constructed in Section~\ref{sec-exp_setting}.
}
We present the major findings of our experiments in Table~\ref{tab:recommandation}.

\begin{figure}[htbp]
    \centering
    \includegraphics[width=1.0\linewidth]{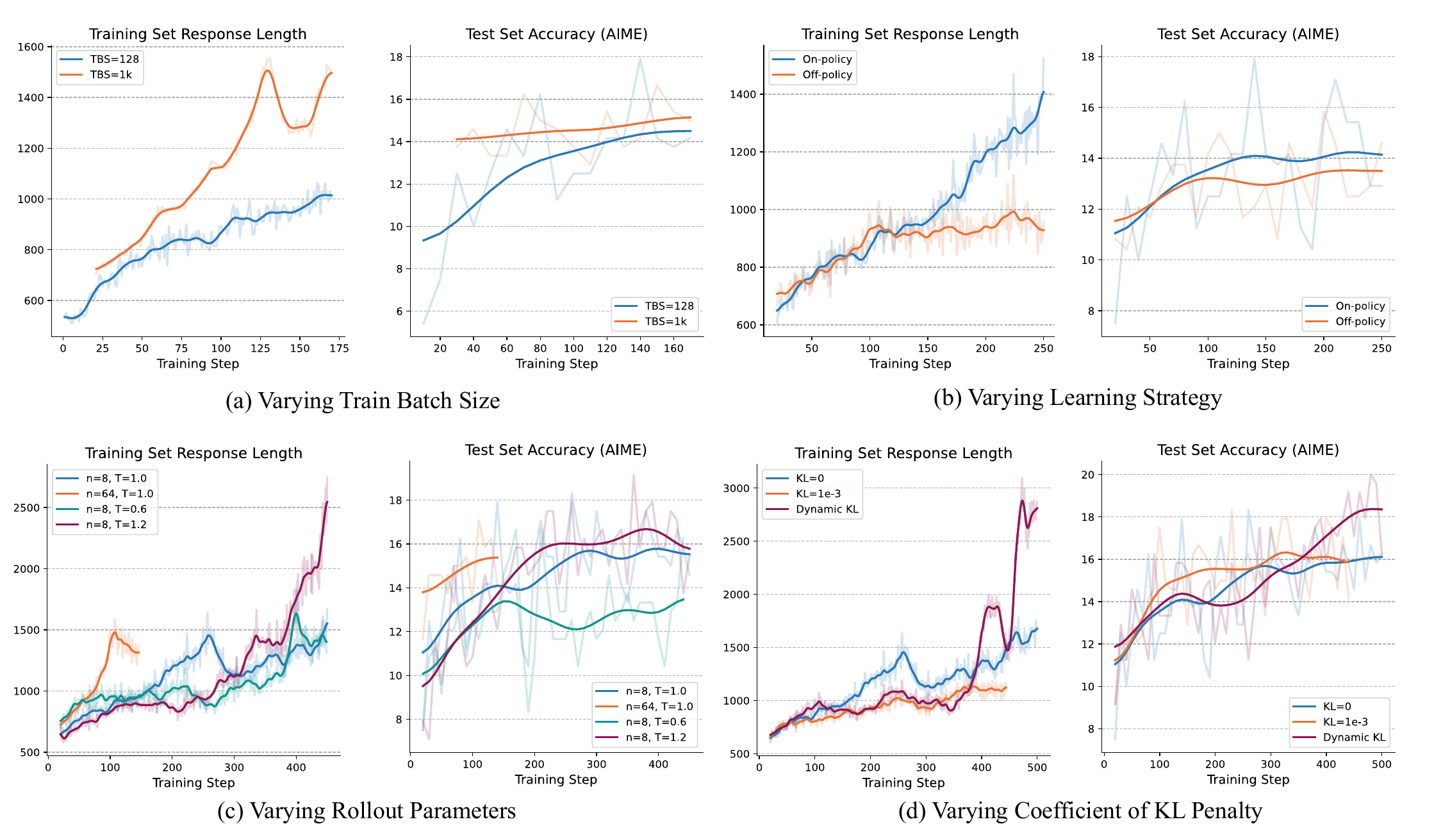} 
    \caption{The performance comparison of \textsc{Qwen2.5-7B} between different RL training settings.}
    \label{fig:para_comp}
\end{figure}


\subsubsection{Influence of Training Hyper-parameters}
\label{sec-hyper_parameters}
In our experiments, we find that certain hyperparameters have significant impact on the stability, learning efficiency, and ultimate performance of RL training. In this part, we present a detailed experimental analysis focusing on hyper-parameter tuning to better understand their effects. 
Considering the training cost, we choose to conduct all of these experiments on \textsc{Qwen2.5-7B}.

\paratitle{Train Batch Size.}
In this report, train batch size~(TBS) refers to the number of queries utilized for training in a single step of the RL process. {A larger value often suggests greater diversity among the queries sampled within that step.}  
In our experiments, we investigate the impact of varying TBS~(\ie $TBS = 128$ \emph{v.s.} $1024$). As illustrated in Figure~\ref{fig:para_comp}~(a), a larger TBS can significantly enhance training efficiency, enabling the model to achieve rapid performance improvements at early training stages. Furthermore, training with larger batch sizes exhibits greater stability, with significantly reduced fluctuations across training metrics compared to smaller batch sizes. 

\paratitle{Learning Strategy: On-policy vs. Off-policy.} 
We primarily explore two learning strategies: on-policy learning and off-policy learning. On-policy learning uses data generated exclusively under the current policy model's distribution for each gradient update. In contrast, off-policy learning utilizes data that may not originate from the current policy. As shown in Figure~\ref{fig:para_comp}~(b), our empirical results demonstrate that on-policy learning yields more favorable outcomes. 
The fully on-policy training approach encourages greater exploration; during training, the model naturally and rapidly increases its response length, whereas off-policy learning with fewer updates struggles with bottlenecks in length growth. Additionally, the on-policy method achieves superior performance on the test set, highlighting the distinct advantages of this strategy.


\paratitle{Rollout Parameters.}
We mainly explore two rollout parameters that affect the model performance: number of rollout times and rollout temperature.
The number of rollout times~($n$) refers to the number of times the policy model samples complete responses for each query during RL process, while the rollout temperature~($T$) denotes the temperature coefficient used during decoding. A larger rollout number and a higher temperature typically indicate a greater degree of exploration. We compare the cases of $n=8$ and $n=64$, and find that increasing the rollout number significantly improves the training performance,  encouraging the model to generate longer responses with more exploration. However, a higher temperature is not always beneficial; elevated temperatures can promote exploration but may also lead to the generation of more meaningless content (repetitions, gibberish, etc.). We conduct comparative experiments with $T$ values of 0.6, 1.0, and 1.2 in Figure~\ref{fig:para_comp}~(c). The results indicate that a lower temperature yielded higher average rewards during the early stages of training, but it might restrict the model's subsequent exploration, resulting in less effective performance in later stages. Conversely, at $T=1.2$, the model exhibits a rapid growth in response length, and outperforms other settings (\ie $T=1.0$ and $T=0.6$) on AIME 2024.
Based on the experiment results, we conclude that a higher temperature tends to result in better performance, in the case that the model can generate text normally without gibberish.


\paratitle{Coefficient of KL Penalty.}
During RL training, the Kullback-Leibler divergence is supposed to impose constraints such that the updated policy does not deviate significantly from the previous behavior policy from which data has been collected.  A larger value of the coefficient of KL penalty results in a stronger constraint, thereby ensuring that the updated policy remains closer to the behavior policy. Hence, the choice of the KL penalty coefficient is critical in balancing the trade-off between exploration and stability in reinforcement learning. We conduct three experiments concerning the KL coefficient: one with no KL penalty ($KL = 0$), one with a fixed KL value ($KL=1\times 10^{-3}$), and another with dynamic KL annealing~(cosine decaying from $KL=1\times10^{-3}$ to $KL=0$). 
In Figure~\ref{fig:para_comp}~(d), we can observe that 
the dynamic KL annealing strategy leads to a better performance than the other variants. 
In the early stages of training, the model is prone to crashing and may get stuck in a local optimum. A fixed KL value can constrain parameter updates, preventing model degradation.
However, as training progresses, a large KL value becomes unsuitable for reinforcement learning, limiting further improvements in the model's capabilities.
By gradually relaxing these constraints, dynamic KL annealing enables continuous improvement of the backbone model.


\paratitle{Effect of Backbone Models.} The experiments described above are based on \textsc{Qwen2.5-7B}. 
Additionally, we also conduct experiments on the smaller \textsc{Qwen2.5-1.5B} model, the supervised fine-tuned \textsc{Qwen2.5-7B-Instruct} model, and the math-specific \textsc{Qwen2.5-Math-7B} model. With the aforementioned settings, our experiments reveal that \textsc{Qwen2.5-7B} demonstrates stronger exploration capabilities compared to \textsc{Qwen2.5-1.5B} and follows a similar trend to \textsc{Qwen2.5-7B-Instruct} during RL training.
However, since \textsc{Qwen2.5-Math-7B} is specialized for the mathematical domain, its inherent biases could influence our assessment of RL training effectiveness. As a result, we have opted not to include these results in this report.

\subsubsection{Impact of the Prompt}

To assess the impact of different prompts on the RL training process, we conduct experiments using two base models (\ie \textsc{Qwen2.5-1.5B} and \textsc{Qwen2.5-7B}) with two types of prompts.  
The first is \emph{short prompt}, similar to the one used in DeepSeek-R1-Zero.  
Additionally, to better elicit the reasoning capabilities of the base models, we design a new prompt that includes detailed instructions about the reasoning process, called \emph{long prompt}.   
Compared to the simple prompt mentioned earlier, this new prompt retains the requirement for a specific reasoning format while adding comprehensive descriptions of the reasoning process. This includes strategies that can be applied during reasoning (\eg analyzing questions, summarizing findings) and recommended expressions and vocabulary to use throughout the process (\eg ``\emph{wait}'', ``\emph{alternatively}'').


Based on our findings in Section~\ref{sec-hyper_parameters} about the training hyper-parameters, we set learning rate, train batch size, rollout temperature, and number of rollout times as $1\times 10^{-6}$, $128$, $1.0$, and $8$ in this experiment, and perform on-policy training strategy.
We set the coefficients for the KL penalty and entropy loss to $0.0$, effectively removing constraints on the model. This allows the performance differences resulting from various prompts to be more clearly observed.


\begin{promptbox}[Prompt with Sample Instruction about Reasoning Process (Short Prompt)]{codegray}
\texttt{A conversation between User and Assistant. The user asks a question, and the Assistant solves it. The assistant first thinks about the reasoning process in the mind and then provides the user with the answer. The reasoning process and answer are enclosed within <think> </think> and <answer> </answer> tags, respectively, i.e., <think> reasoning process here </think> <answer> answer here </answer>.\\The assistant shows the reasoning process in <think> </think> tags, and returns the final answer in <answer> </answer> tags, for example <answer> $\frac{1}{2}$ </answer>.\\User: \{QUESTION\}\\Assistant:\\<think>}
\end{promptbox}

\begin{promptbox}[Prompt with Detailed Instruction about Reasoning Process (Long Prompt)]{codegray}
\texttt{A conversation between User and Assistant. The User asks a question, and the Assistant solves it. The Assistant first engages in an internal reasoning process, akin to a stream of consciousness, before providing the User with the answer. The reasoning process and answer are enclosed within `<think></think>' and `<answer></answer>' tags, respectively. For example:\\\\```\\<think>\\reasoning process here\\</think>\\<answer>\\answer here\\</answer>\\'''\\\\The reasoning process includs detailed considerations such as analyzing questions, summarizing relevant findings, brainstorming new ideas, verifying the accuracy of current steps, refining any errors, and revisiting previous steps. During this process, the Assistant uses casual, genuine phrases such as: "Hmm", "Wait", "Alternatively", "double check", "I wonder...", "But", "rethink", etc., to make the reasoning process coherent, clear, and logically sound, effectively simulating human cognitive processes.\\\\The Assistant shows the reasoning process within `<think></think>' tags, and ONLY return the FINAL ANSWER within `<answer></answer>' tags. For example: `<answer> $\frac{1}{2}$ </answer>'.\\\\User: \{QUESTION\}\\Assistant:\\<think>\\}
\end{promptbox}

\begin{figure}[htbp]
    \centering
    \includegraphics[width=1.0\linewidth]{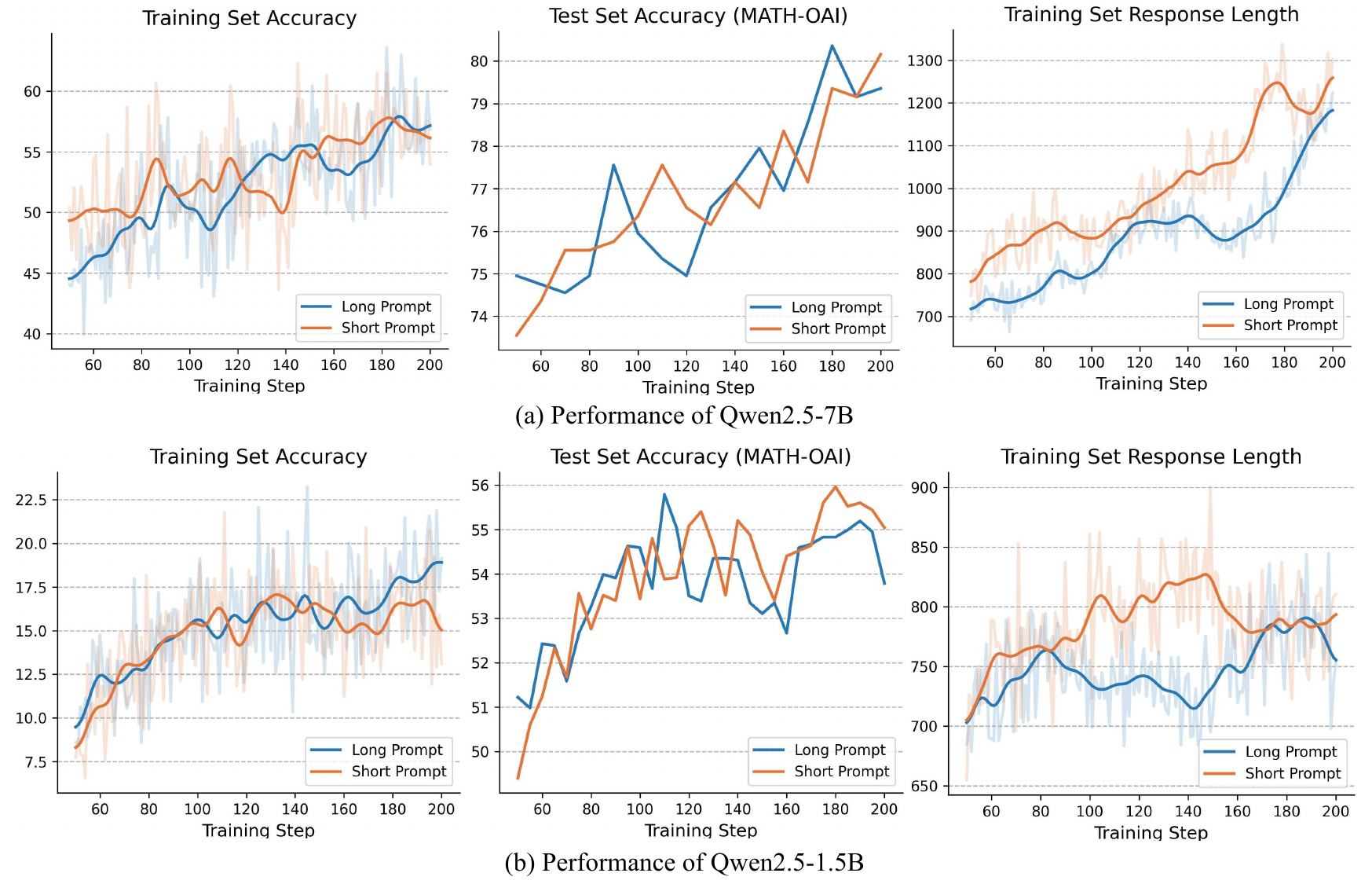} 
    \caption{The accuracy on training set and test set, and the average length of responses of \textsc{Qwen2.5-7B} and \textsc{Qwen2.5-1.5B} trained on different prompts.}
    \label{fig:long_short_prompt}
\end{figure}

The experimental results are presented in Figure~\ref{fig:long_short_prompt}.  
From this figure, we observe that \textsc{Qwen2.5-1.5B} and \textsc{Qwen2.5-7B} exhibit different behaviors when trained on the two types of prompts.  
For \textsc{Qwen2.5-1.5B}, the model trained on the short prompt achieves higher performance on the test set compared to the model trained on the long prompt.  
This may be because a 1.5B-sized base model has relatively limited capacity and struggles to follow the complex instructions in the detailed prompts.  
In contrast, the 7B-sized model shows similar performance on downstream tasks when trained on different prompts. However, the model trained on the long prompt generates shorter responses, suggesting that it learns to reason more efficiently by adhering to the guidelines provided in the prompt.  
Thus, we conclude that more detailed prompts can guide the model to think more effectively, enhancing reasoning efficiency. However, they do not necessarily lead to improved performance on downstream tasks. 





\subsection{\textsc{STILL-3-Zero-32B}: RL Improves Reasoning Ability on Base Model}
\label{sec-still_3_zero}

In this section, we aim to examine how base models can be elevated into a high level of complex reasoning by RL training.
Following our previous study, we adopt the carefully-tuned settings for training, including the selection of hyper-parameters, backbone models, and the training data.

\paratitle{RL Setting.}
As demonstrated in our earlier experiments (Section~\ref{sec:rl_setting}), we observe that smaller-sized base models often tend to be less effective in  possessing complex reasoning capacities. Therefore, in this section, we adopt \textsc{Qwen2.5-32B} as the backbone model for further experiments. 
Considering both the effectiveness and the consumption of computational resources, we set the training batch size as $128$, and rollout $16$ times, performing an on-policy and efficient training process.
We adopt $1\times 10^{-6}$ as the learning rate and turn off the KL penalty to remove the constraints imposed by the reference model, {and utilize the 90k training questions (Section~\ref{sec-exp_setting}).}
Moreover, to conduct a context-efficient training process, we first set the maximum context length to 8k and then gradually extend it to 20k.
Since the base model tends to generate excessively long and meaningless text in the early stages of training (\eg repetitive text or gibberish), setting a small context length can early stop such meaningless content to improve training efficiency.

\begin{figure}[htbp]
    \centering
    \includegraphics[width=1.0\linewidth]{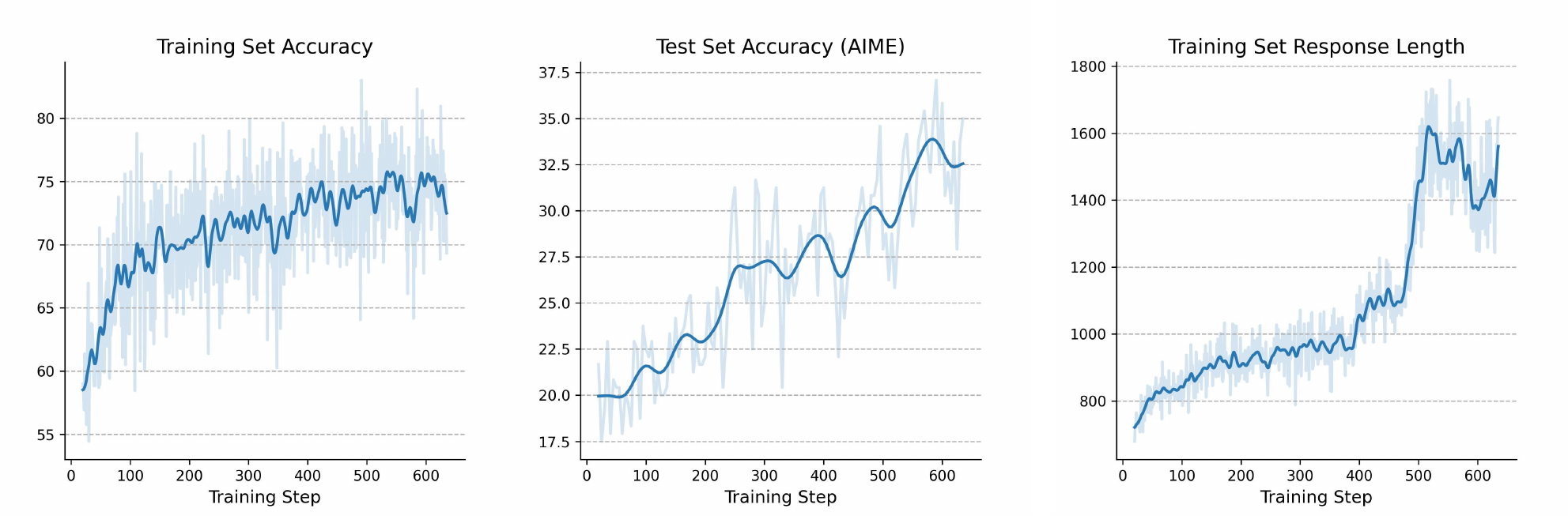} 
    \caption{The accuracy on training set and test set, and the average length of responses of \textsc{STILL-3-Zero-32B} during RL process.}
    \label{fig:still_3_zero_32b}
\end{figure}

\paratitle{Performance of \textsc{STILL-3-Zero-32B}.}
We present the performance of \textsc{STILL-3-Zero-32B} in Figure~\ref{fig:still_3_zero_32b}.
At the beginning of the training process, the response length of the model is around 500 tokens and gradually increases.
We can find that the 8k context length of early stage RL process will not affect the model's ability to learn reasoning skills or increase response length.
During the training process, the model's accuracy on AIME 2024 increased from 2.08\% to 37.08\%, accompanied by an increase in response length, indicating that the model has learned to spend more time on thinking before obtaining the final answer.
Analyzing the performance curve of the model, we find that it is still in a growth phase.
This phenomenon has shown that scaling the training steps of RL can bring a larger improvement on the reasoning ability of the backbone model.
Moreover, to better demonstrate the effect of RL process, we present a comparison of the responses generated by \textsc{Qwen2.5-32B} and \textsc{STILL-3-Zero-32B} in Table~\ref{case_study_qa}.
We can observe that the complex reasoning actions (\eg reflection and verification) have emerged after RL training, and the model tends to engage in more thorough reasoning before providing the final answer.

\begin{figure}[htbp]
    \centering
    \includegraphics[width=0.64 \linewidth]{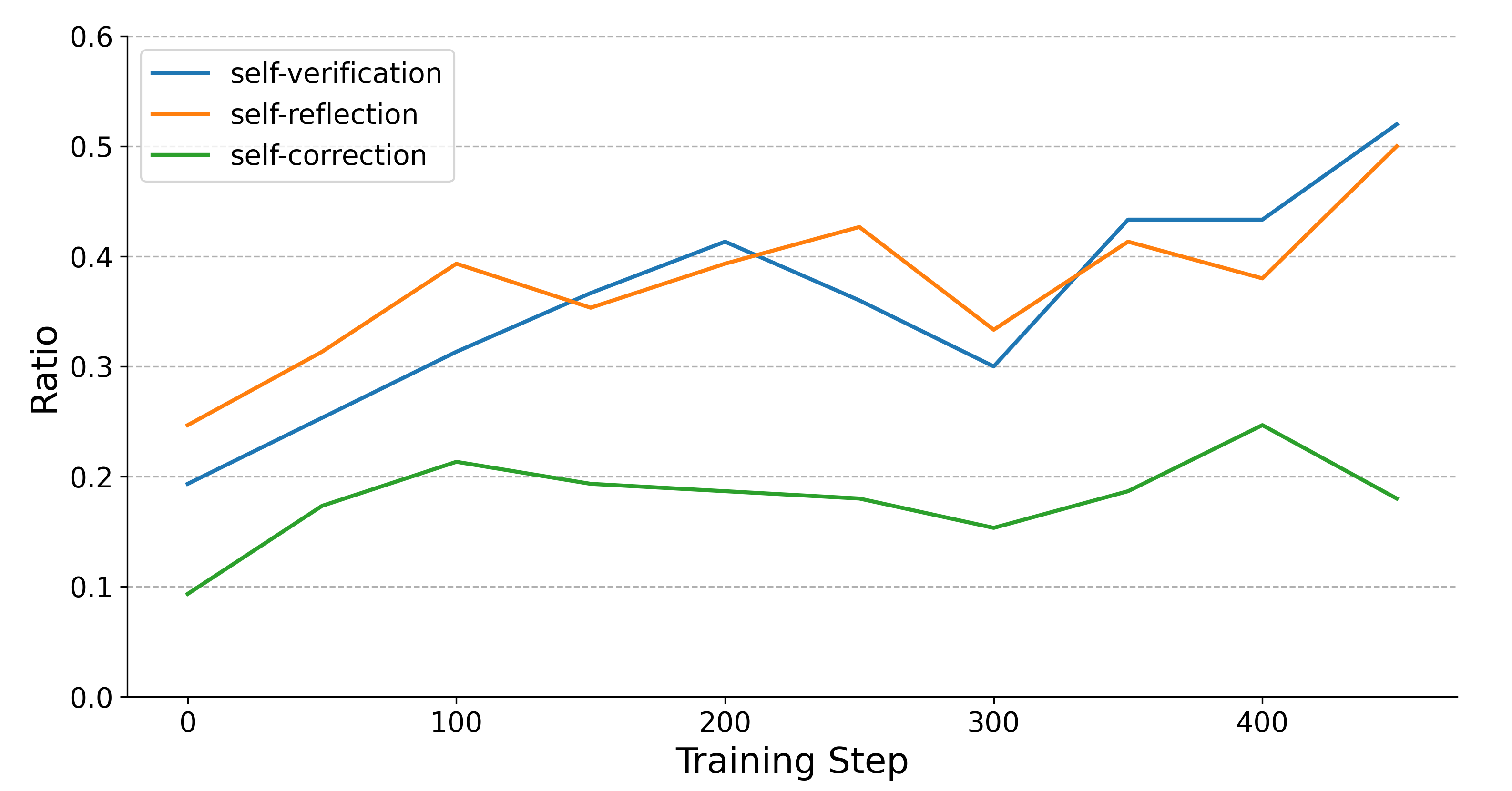} 
    \caption{The ratio of keyword categories in responses generated by \textsc{STILL-3-Zero-32B} on AIME 2024. The self-verification category includes words like ``\emph{verify}'', ``\emph{double check}'', and ``\emph{confirm}''. The self-reflection category contains terms such as ``\emph{however}'', ``\emph{reflect}'', and ``\emph{wait}''. The self-correction category encompasses words including ``\emph{correct}'', ``\emph{revise}'', and ``\emph{adjust}''.}
    \label{fig:aha_moment}
\end{figure}

\begin{table*}[ht]
    \small
    \centering
    \begin{tabular}{ll}
        \toprule
        \multicolumn{1}{l}{\begin{tabularx}{0.1\textwidth}{@{}X@{}}
            \textbf{Problem}
        \end{tabularx}} & \multicolumn{1}{l}{\begin{tabularx}{0.8\textwidth}{@{}X@{}}
            Let $a$ be a positive real number such that all the roots of \[x^3 + ax^2 + ax + 1 = 0\]are real.  Find the smallest possible value of $a.$
        \end{tabularx}} \\
        \midrule
        \multicolumn{1}{l}{\begin{tabularx}{0.1\textwidth}{@{}X@{}}
            \textbf{\textsc{Qwen2.5-32B}}
        \end{tabularx}} & \multicolumn{1}{l}{\begin{tabularx}{0.8\textwidth}{@{}X@{}}
            <think>First, let's find the roots of the given cubic equation. We can use the rational root theorem to check for possible rational roots. The possible rational roots are $\pm 1$. However, neither $1$ nor $-1$ is a root of the equation. Therefore, there are no rational roots. Next, we can use the discriminant of the cubic equation to determine if there are real roots. The discriminant of a cubic equation $ax^3 + bx^2 + cx + d = 0$ is given by $D = 18abcd - 4b^3d + b^2c^2 - 4ac^3 - 27a^2d^2$. If $D > 0$, there are three distinct real roots; if $D = 0$, there are at least two equal real roots; and if $D < 0$, there is one real root and two complex roots. In our case, $a = 1$, $b = a$, $c = a$, and $d = 1$. So, the discriminant is $D = 18a - 4a^3 + a^4 - 4a^3 - 27$. To ensure that all roots are real, we need $D \geq 0$. Now, let's find the minimum value of a that satisfies this condition.</think>\\<answer>2</answer>
        \end{tabularx}} \\
        \midrule
        \multicolumn{1}{l}{\begin{tabularx}{0.1\textwidth}{@{}X@{}}
            \textbf{\textsc{STILL-3-Zero-32B}}
        \end{tabularx}} & \multicolumn{1}{l}{\begin{tabularx}{0.8\textwidth}{@{}X@{}}
            <think>\\To determine the smallest possible value of \( a \) such that all the roots of the polynomial \( x^3 + ax^2 + ax + 1 = 0 \) are real, we need to analyze the properties and constraints of the roots.\\... \\\textcolor{red}{We need to find the smallest possible value of \( a \).} ... \\Now, let's consider the case of one negative root and two positive roots. Let \( r = -u \) (negative root) and \( s = v \), \( t = w \) (positive roots).\\... \\We need to find a consistent set of \( u, v, w \) that satisfy these equations. \textcolor{red}{However, the analysis of this case is more complex}, and we already have a valid solution from the case of three negative roots.\\\\\textcolor{red}{Let's double-check if there could be a smaller value of \( a \) than 3}. From the AM-GM inequality, we know that the minimum value of \( u + v + w \) and \( uv + uw + vw \) is 3 when \( u = v = w = 1 \). Any deviation from \( u = v = w = 1 \) would result in a value of \( a \) that is not necessarily smaller than 3 and would need to satisfy the polynomial having all real roots.\\\\Thus, the smallest possible value of \( a \) that satisfies the condition that all the roots of the polynomial \( x^3 + ax^2 + ax + 1 = 0 \) are real is \( a = 3 \).\\</think>\\<answer>3</answer>
        \end{tabularx}} \\
        \bottomrule
    \end{tabular}
    \caption{The comparison of the responses from \textsc{Qwen2.5-32B} and \textsc{STILL-3-Zero-32B}.}
    \label{case_study_qa}
\end{table*}

\paratitle{Aha Moment: Emergence of the Complex Reasoning Patterns.}
The Aha Moment in RL training refers to sudden qualitative leaps in reasoning patterns~\cite{deepseek-r1}, where models spontaneously develop behavioral patterns that exhibit superficial similarities to the strategies humans employ when solving problems.
In our experiments, we identify its emergence through statistical analysis of the keyword frequency of complex reasoning actions during RL training.
Specifically, we manually curate a set of reasoning-indicating keywords that reflect human-like reasoning patterns, such as reflection, verification, and correction. We then track the changes in the proportion of these keywords in the responses generated by the model during RL training.  
As illustrated in Figure~\ref{fig:aha_moment}, during the training of \textsc{STILL-3-Zero-32B}, the model begins to exhibit reasoning behaviors at an early phase. For example, even at training step 0, the ratio calculated from the base model's evaluation on AIME 2024 is approximately 0.1, indicating that these reasoning actions are already inherent in the base model. As the RL training progresses, these patterns are further strengthened, indicating that RL continuously prompts the model to better utilize and enhance its reasoning skills, making the complex reasoning patterns more distinct.

\section{Experiments on Fine-tuned Models}
\label{sec-code-start-model}

After examining ``Zero'' RL effect, we continue to study how to improve the complex reasoning capacities of fine-tuned language models. 

\subsection{Fine-tuning Base Models with Long CoT Data}
\label{sec-finetuned_longcot_data}

Although the base model demonstrates promising results during training, we aim to further investigate whether incorporating a small amount of high-quality data as cold start data can improve RL performance. For this purpose, we first explore two distinct data synthesis methods and subsequently compare their effects after RL training. 

\subsubsection{Long CoT Training Data Construction}
There are two primary methods for constructing long CoT training data: synthesizing long CoT data and distilling from large reasoning models. For the first method, we synthesize instruction tuning data that incorporates various reasoning patterns~(\eg reflection and verification) to encourage exploration and diversify reasoning patterns. {Following existing work~\cite{Hou-arxiv-2025-Advancing}, we select approximately 5k hard problems from their public dataset}. For the second method, we perform distillation using existing large reasoning models, as outlined in prior studies~\cite{still2}. We select around 50k data from OpenThoughts~\cite{openthoughts}, which distills long CoT responses from \textsc{DeepSeek-R1}. We then conduct post-processing to select responses of appropriate length and accuracy. After preparing these two types of long CoT training data, we apply a cold start to the base model with supervised fine-tuning, respectively.

\begin{table}[htbp]
    \centering
    \small
    \setlength\tabcolsep{2.4pt}
    \caption{Performance of \textsc{Qwen2.5-7B} trained on different reasoning instruction data.}
      \begin{tabular}{lcccccccc}
      \toprule
      \multirow{2.5}*{\textbf{Methods}} & \multicolumn{2}{c}{\textbf{MATH-OAI}} & \multicolumn{2}{c}{\textbf{AIME 2024}} & \multicolumn{2}{c}{\textbf{OMNI}} & \multicolumn{2}{c}{\textbf{LiveAOPS}} \\
      \cmidrule(r){2-3}\cmidrule(r){4-5}\cmidrule(r){6-7}\cmidrule(r){8-9}
      & Accuracy & Length & Accuracy & Length & Accuracy & Length & Accuracy & Length \\
      \midrule
      Backbone & 47.04 & 630 & 3.33 & 2202 & 19.00 & 1430 & 17.34 & 1601 \\
      Synthesis Data (5k) & 19.72 & 1151 & 6.67 & 2254 & 5.80 & 1917 & 10.21 & 2764 \\
      Distillation Data (50k) & 82.88 & 5078 & 33.33 & 15507 & 38.40 & 9380 & 49.52 & 9057 \\
      \bottomrule
      \end{tabular}
      \label{tab:sft_results}
\end{table}

\begin{figure}[htbp]
    \centering
    \includegraphics[width=1.0\linewidth]{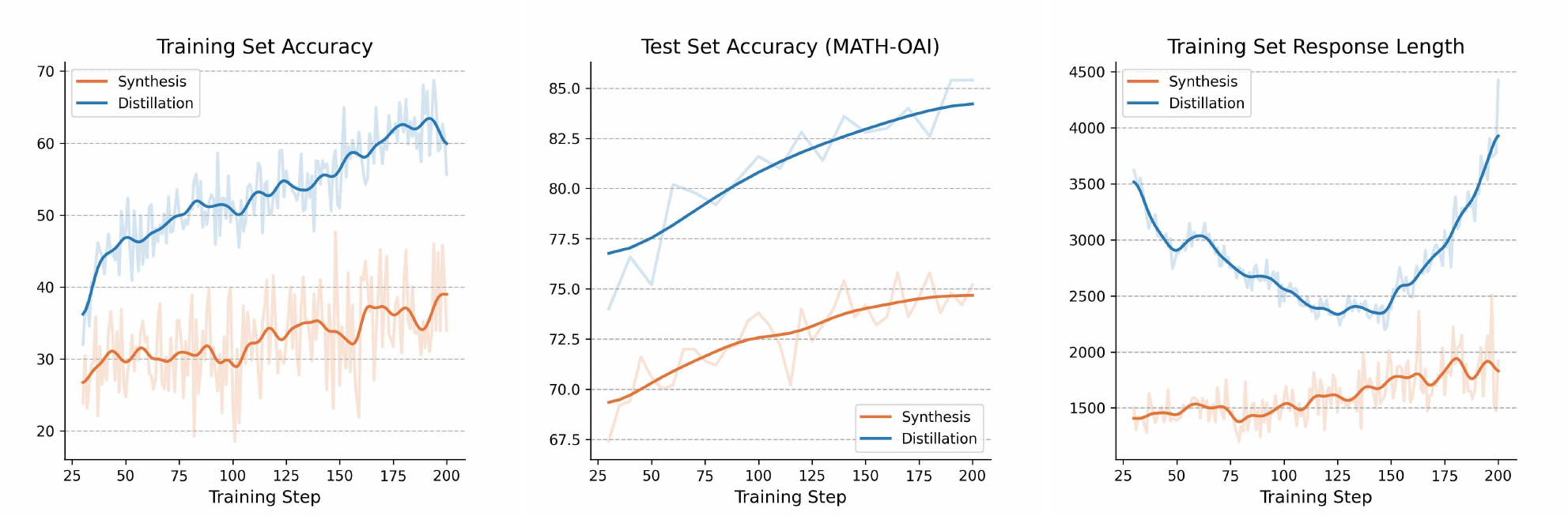} 
    \caption{The accuracy on training set, the average length of responses, and the ratio of completed responses of \textsc{Qwen2.5-7B} fine-tuned on different reasoning data during the RL process.}
    \label{fig:reasoning_data}
\end{figure}

\subsubsection{Evaluation and Analysis}
In the following, we present the experimental results by either SFT or RL. 

\paratitle{Models with SFT Training.}
After fine-tuning the backbone model with the above CoT instruction data, we evaluate its performance and reasoning length on the MATH-OAI, AIME 2024, OMNI, and LiveAOPS benchmarks mentioned in Section~\ref{sec-exp_setting}, and present the evaluation results in Table~\ref{tab:sft_results}. The results indicate that both the synthesis and distillation methods encourage the model to generate longer responses, with the distillation method significantly enhancing response length compared to that of the synthesis method. Additionally, the synthesis method struggles to effectively elicit the reasoning capacities of the backbone model, whereas the distillation method consistently improves overall performance relative to the backbone model. This is primarily because synthesized data is typically generated using relatively simple strategies, which are insufficient to capture or represent the complex reasoning procedures that are more effectively distilled from highly capable reasoning models.

\paratitle{Models with RL Training.}
After performing cold-start initialization using the two aforementioned methods, we conduct RL training under consistent experimental settings to compare their performance (refer to Section~\ref{sec:rl_setting} for RL settings). The results are presented in Figure~\ref{fig:reasoning_data}.   
For the model initialized by synthesizing long-chain data from scratch, the initial sequence length is relatively short. As a result, during training, the sequence length steadily increases alongside improvements in accuracy. However, due to the limited quality of the synthesized long CoT data and the model's initially poor performance, more training steps are required to achieve results comparable to the distillation-based variant.  
In contrast, the model initialized through distillation from an existing LRM exhibits a longer initial sequence length and better initial performance. During subsequent RL training, while accuracy continues to improve, the sequence length may fluctuate or even decrease. Nevertheless, the higher initial performance allows this model to achieve superior results more quickly.

\begin{table}[htbp]
    \centering
    \small
    \setlength\tabcolsep{2.4pt}
    \caption{The length of correct responses and incorrect responses of different large reasoning models.}
      \begin{tabular}{lcccccc}
      \toprule
      \multirow{2.5}*{\textbf{Models}} & \multicolumn{3}{c}{\textbf{MATH-OAI}} & \multicolumn{3}{c}{\textbf{AIME 2024}}\\
      \cmidrule(r){2-4}\cmidrule(r){5-7}
      & Accuracy & Correct & Incorrect & Accuracy & Correct & Incorrect \\
      \midrule
      \textsc{DeepSeek-R1-Distill-Qwen-1.5B} & 84.04 & 3791.70 & 10484.63 & 28.67 & 6987.25 & 13400.67 \\
      \textsc{DeepSeek-R1-Distill-Qwen-7B} & 92.66 & 3295.42 & 14442.13 & 52.00 & 7634.31 & 19419.58 \\
      \textsc{DeepSeek-R1-Distill-Qwen-32B} & 94.88 & 3619.36 & 10720.28 & 70.42 & 8058.97 & 14695.77 \\
      \bottomrule
      \end{tabular}
      \label{tab:response_length}
\end{table}

\begin{figure}[htbp]
    \centering
    \includegraphics[width=1.0\linewidth]{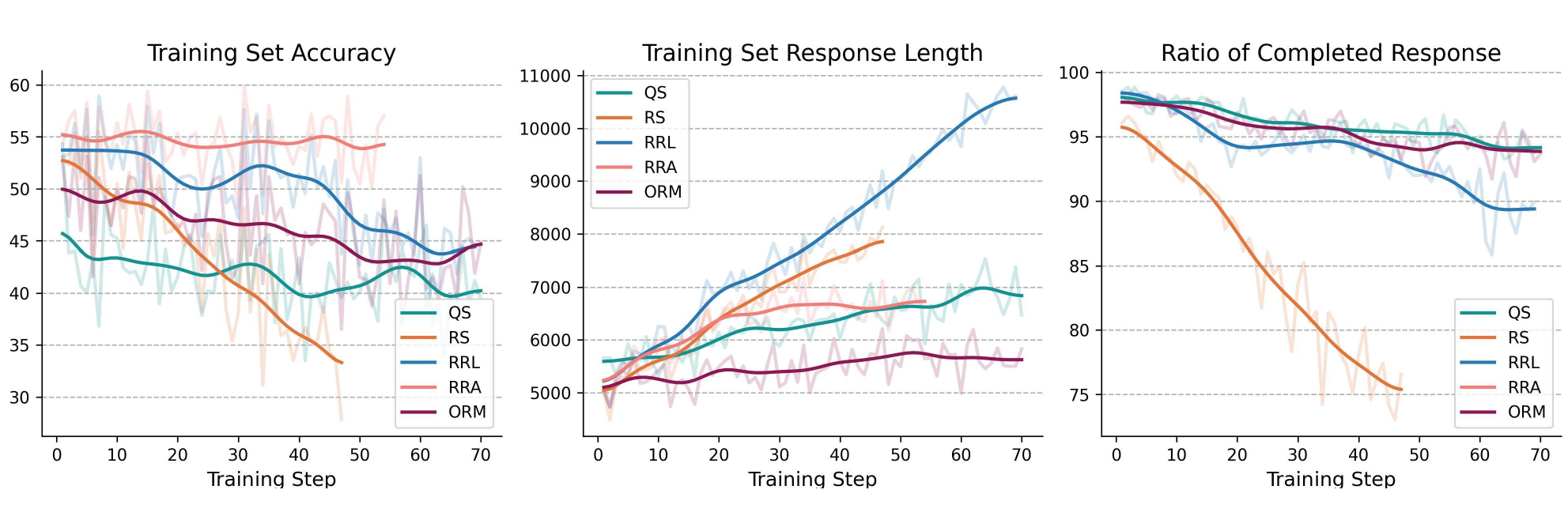} 
    \caption{The accuracy on training set, the average length of responses, and the ratio of completed responses of the model trained on different auxiliary approaches during the training process.}
    \label{fig:auxiliary_approach}
\end{figure}

\subsection{Discussion about Length Hacking in RL}
\label{sec:Auxiliary Approaches in RL}

As highlighted in prior work~\cite{deepseekr1,kimik1.5}, longer response lengths often correlate with a larger search space, which can enhance the reasoning capabilities of LLMs. However, we empirically observe that LLMs can exploit length-biased reward functions (\ie reward hacking), which are designed to encourage longer responses and more complex reasoning behaviors. In the following sections, we will first discuss our length encouragement strategies, and then present the results and analysis. 



\subsubsection{Length Encouragement Strategies} 
We consider the following length encouragement strategies: 

\paratitle{Question Selection (QS).} 
In our previous evaluation in Table~\ref{tab:response_length}, we find that the average length of correct responses is much shorter than the average length of incorrect responses.
At the same time, the correct responses will receive the higher rewards than the incorrect responses.
Based on this phenomenon, the LLMs may learn from the training data that generating shorter responses leads to higher scores, which violates the goal of the RL process.
To prevent this situation, we calculate the average length of correct and incorrect responses, \ie $L_{c}$ and $L_{w}$, and only choose the questions that satisfy $L_{c} > L_{w}$ to perform advantage estimation and gradient descent.

\paratitle{Response Selection (RS).}
Similar with the question selection strategy mentioned above, the major goal of response selection is to guide the LLMs to learn to generate longer responses, which can extending the search space during the reasoning process.
During the RL process, after obtaining the generated solution for the question, we check its correctness and calculate the length of each solution. 
If the length of the longest correct solution exceeds that of the shortest incorrect solution, we guide the model to learn from both these two solutions.
In this way, the model can learn to increase the length of the generated solutions.

\paratitle{Rewards on Response Length (RRL).}
Although question selection and response selection strategies can teach LLMs to generate a longer solution, these methods train the model using only a small portion of the rollout instances, causing a waste of data and computational resources.
We design a new reward function that assigns higher reward scores for longer responses, in order to avoid data waste.
Formally, the reward function can be defined as:
$$R_{\text{Length}}(i) = \frac{L_i}{L_\text{Max}} + R_{\text{Correctness}}(i),$$ where $L_i$ and $L_{\text{Max}}$ denote the length of $i$-th response and the the maximum context window length, and $R_{\text{Correctness}}(i)$ is discrete function that returns $1$ for correct solution while $-1$ for incorrect solution.
In this reward function, the correct responses will receive higher rewards than the incorrect ones, and a longer response will receive higher rewards than a shorter response no matter whether they are correct or incorrect.

\paratitle{Rewards on Reasoning Action (RRA).}
To encourage LLMs to perform complex reasoning actions, we also design a special reward function, providing higher rewards for the solutions containing more complex reasoning actions.
Similar with the length reward $R_{\text{Length}}(i)$, the reward of reasoning action can be defined as: $$R_{\text{Action}}(i)=\frac{\min(A_i, A_\text{Max})}{A_\text{Max}} + R_{\text{Correctness}}(i),$$ 
where $A_i$ denotes the number of reasoning actions in the generated solution, and $A_\text{Max}$ is a preset upper bound, designed to prevent excessively high rewards and mitigate the issue of reward hacking.
In practice, we calculate the word frequency of keywords (\ie ``\emph{however}'', ``\emph{but}'', ``\emph{wait}'', ``\emph{verify}'', and ``\emph{alternatively}'') in the model's generated responses and used this frequency as the value for $A_i$.
Additionally, we set $A_\text{Max}$ as $20$ to prevent the model only learning to generate the keywords.

\paratitle{Overlength Response Masking (ORM).} 
During RL training, we truncat some responses before completion due to the limited maximum window length. This truncation often results in the inability to extract the final answer. 
In this case, a potential methods is to consider these overlength responses as the incorrect solutions and assign them negative rewards.
However, determining whether these truncated responses are inherently incorrect is challenging, as the incompleteness may not reflect the actual quality of the reasoning or content. To address this issue, we exclude these incomplete responses from the training process and focus solely on fully completed responses for RL training. This approach enhances the precision of reward scoring and ensures a more reliable training signal. 

\subsubsection{Empirical Results and Analysis} 

To carry out our experiments, we adopt the similar RL training method, while incorporate the aforementioned length encouragement strategies. 
during the training process, we monitor changes in performance and response length, as illustrated in Figure~\ref{fig:auxiliary_approach}. 
Based on the experimental results, we observe that each of the aforementioned approaches gradually increases the response length of the backbone model during the RL process.  
The reward RRL is more effective at extending response length compared to other approaches, while the impact of the overlength response masking strategy is relatively limited.  
However, as response length increases, the model's performance appears to be significantly affected across all variants except RRA. Unlike the others, RRA does not prioritize longer responses but instead focuses on encouraging more reasoning actions.
Upon manually inspecting the generated samples, we find that the ratio of completed responses tends to decrease as response length grows.  
Specifically, the ratio of completed responses measures the proportion of responses that the model is able to fully generate without exceeding the maximum context length.
A lower completion ratio indicates that more questions remain unanswered, resulting in a decline in accuracy. This serves as a critical quality indicator for monitoring solutions during RL training. 


These results indicate that explicitly training LLMs to extend their thought processes or response length may unintentionally degrade model performance, leading to significant reward hacking. 
Instead, length extension should arise naturally through self-exploration, driven by the model's intrinsic learning processes rather than being directly encouraged by rewards.



\begin{figure}[htbp]
    \centering
    \includegraphics[width=1.0\linewidth]{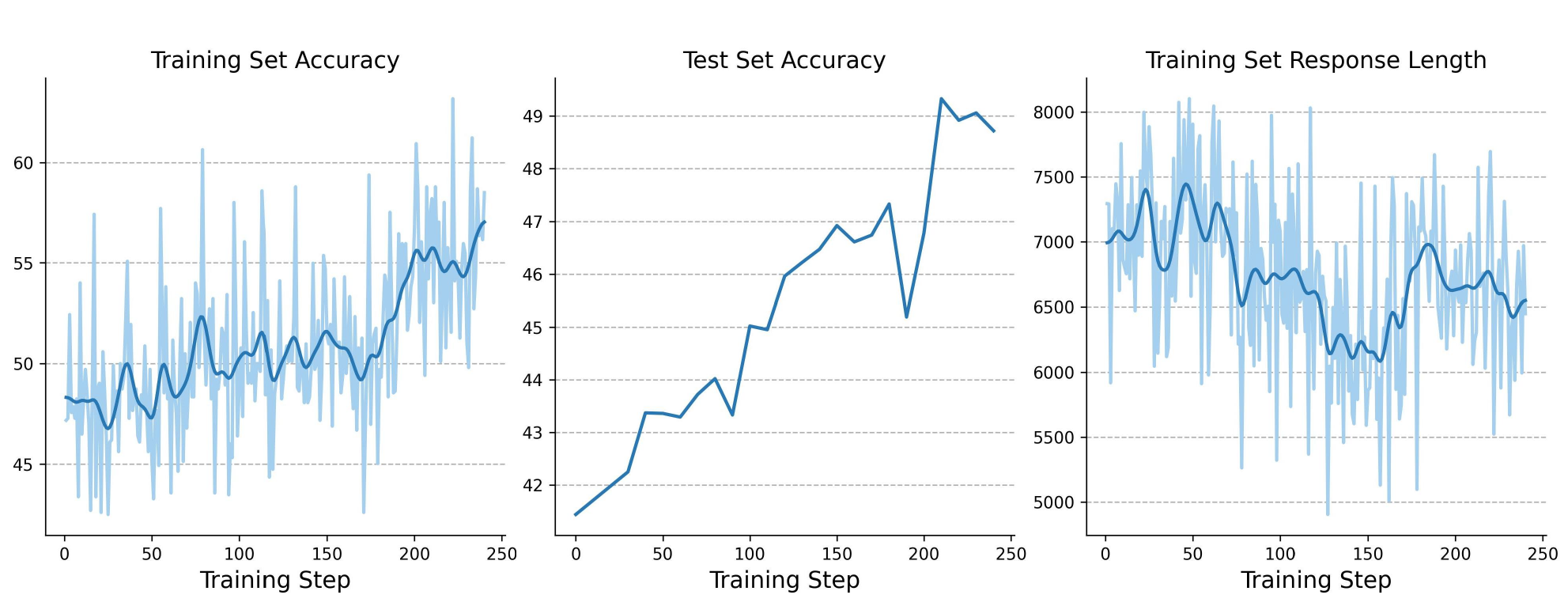} 
    \caption{The accuracy on training set and test set, and the average length of responses of \textsc{STILL-3-1.5B} during the training process.}
    \label{fig:still_3_1.5B}
\end{figure}

\begin{table}[htbp]
    \centering
    \small
    \setlength\tabcolsep{2.4pt}
    \caption{Performance of \textsc{STILL-3-1.5B} and its backbone model \textsc{DeepSeek-R1-Distill-Qwen-1.5B}. For MATH-OAI and AIME 2024, we sample 5 responses and calculate the Pass@1 accuracy. For OMNI and LiveAOPS, we utilize greedy decoding strategy for evaluation.}
      \begin{tabular}{lccccc}
      \toprule
      \textbf{Models} & \textbf{MATH-OAI} & \textbf{AIME 2024} & \textbf{OMNI} & \textbf{LiveAOPS} & \textbf{Avg.} \\
      \midrule
      \textsc{DeepSeek-R1-Distill-Qwen-1.5B} & 84.04 & 28.67 & 25.60 & 33.33 & 42.91 \\
      \textsc{STILL-3-1.5B} & 85.48 & 39.33 & 33.00 & 39.50 & 49.33 \\
      \bottomrule
      \end{tabular}
      \label{tab:still_3_1.5B}
\end{table}



\subsection{\textsc{STILL-3-1.5B}: Enhancing Slow Thinking Abilities of Small Models via RL}
\label{sec-still-3-1.5}

In this section, we investigate how small models can achieve advanced levels of complex reasoning through RL training. Rather than applying RL directly to base or instruction-tuned models, we focus on small models that have been trained for slow, deliberate reasoning. This approach is motivated by the observation that small models, due to their limited capacity, may not effectively explore and learn under the guidance of rule-based rewards alone. Our objective is to explore whether RL can further enhance the reasoning capabilities of small reasoning models that have undergone specialized long CoT instruction tuning. By doing so, we aim to evaluate the potential advantages of RL training in augmenting the performance of such models.


\paratitle{RL Settings.}  
To conduct our experiments, we select \textsc{DeepSeek-R1-Distill-Qwen-1.5B} as the target small model for improvement. As demonstrated in the DeepSeek-R1 paper~\cite{deepseekr1}, this distilled small model has already achieved a high level of performance in mathematical reasoning tasks. For instance, it attains a score of 28.9 (pass@1) on the test set of AIME 2024. A key objective of our study is to investigate whether this performance represents the upper limit of small models or if further enhancements are possible. 
For model optimization, we employ the following hyperparameters:  a  learning rate of $2\times10^{-6}$, a training batch size of $128$, a mini-batch size as $64$, and 
8 rollouts per training step. These settings are designed to facilitate effective RL training while maintaining computational efficiency.
To prevent the model from crashing, we set the coefficient of the KL penalty as  $1\times10^{-3}$. Additionally, we periodically replace the reference model after a predefined number of training steps. This strategy not only alleviates constraints on the model’s exploration but also facilitates further performance improvements.
{Besides, we train the model with a chat template as shown in the top part of Table~\ref{tab:still_3_1.5b_prompt}} and do not utilize system prompt or other instructions.

\begin{table*}[ht]
    \small
    \centering
    \caption{The prompt template utilized in STILL-3-1.5B RL process and the influence of different prompt templates on  AIME 2024 during evaluation process.}
    \begin{tabular}{lc}
        \toprule
        \multicolumn{1}{l}{\begin{tabularx}{0.6\textwidth}{@{}X@{}}
            \textbf{Prompt Template}
        \end{tabularx}} & \multicolumn{1}{c}{\begin{tabularx}{0.09\textwidth}{@{}X@{}}\centering
            \textbf{Accuracy}
        \end{tabularx}} \\
        \midrule
        \multicolumn{2}{c}{\textit{Training Prompt}} \\
        \\
        \multicolumn{1}{l}{\begin{tabularx}{0.6\textwidth}{@{}X@{}}
            <|begin\_of\_sentence|><|User|>\{QUESTION\}<|Assistant|>
        \end{tabularx}} & \multicolumn{1}{c}{\begin{tabularx}{0.09\textwidth}{@{}X@{}}\centering
            -
        \end{tabularx}} \\
        \midrule
        \midrule
        \multicolumn{2}{c}{\textit{Testing Prompt}} \\
        \\
        \multicolumn{1}{l}{\begin{tabularx}{0.6\textwidth}{@{}X@{}}
            <|begin\_of\_sentence|><|User|>\{QUESTION\}<|Assistant|>
        \end{tabularx}} & \multicolumn{1}{c}{\begin{tabularx}{0.09\textwidth}{@{}X@{}}\centering
            39.33
        \end{tabularx}} \\
        \midrule
        \multicolumn{1}{l}{\begin{tabularx}{0.6\textwidth}{@{}X@{}}
            <|begin\_of\_sentence|><|User|>\{QUESTION\}\\Please reason step by step, and put your final answer within boxed\{\}.<|Assistant|>
        \end{tabularx}} & \multicolumn{1}{c}{\begin{tabularx}{0.09\textwidth}{@{}X@{}}\centering
            33.33
        \end{tabularx}} \\
        \bottomrule
    \end{tabular}
    \label{tab:still_3_1.5b_prompt}
\end{table*}

\paratitle{Performance of \textsc{STILL-3-1.5B}.} 
Figure~\ref{fig:still_3_1.5B} illustrates the trends in training set accuracy,  test set accuracy, and  response length of the training set throughout the training process. Additionally, we provide the final results for \textsc{STILL-3-1.5B} and its backbone model in Table~\ref{tab:still_3_1.5B}.  
The evaluation results show that \textsc{STILL-3-1.5B} achieves significant improvements across all downstream tasks, notably reaching a 39.33\% accuracy on the AIME task, which represents a relative improvement of 37.18\%. 
Compared to the performance of \textsc{Qwen2.5-1.5B} trained with RL mentioned in Section~\ref{sec-hyper_parameters}, the backbone model is specifically trained on distilled high-quality long CoT data, enabling the smaller model to develop robust reasoning capabilities. 
Unlike RL, SFT relies on imitation learning, which is particularly beneficial for small-sized models that typically have limited exploration capacity.
Moreover, the consistent increase in training set accuracy throughout the training process indicates that the model progressively refined its reasoning capabilities.   
Regarding response length, we observe a slight reduction in the length of generated responses during training. 
As discussed in Section~\ref{sec-code-start-model}, this reduction may be attributed to the elimination of redundant reasoning steps initially imitated from the distillation dataset, given that the original response length was notably high. These results demonstrate that RL can also significantly enhance the reasoning capabilities of a well-tuned small reasoning model. 
Besides, during our evaluation, we observe that incorporating additional prompts significantly impacts the model's performance. Table~\ref{tab:still_3_1.5b_prompt} compares the performance of \textsc{STILL-3-1.5B} with and without additional prompts: using prompts such as ``\emph{reason step by step}'' results in an accuracy drop of 6 points.  
In alignment with OpenAI's guidelines for reasoning models, we avoid using specific prompts to elicit responses from the model.

\begin{table}[htbp]
    \centering
    \small
    \setlength\tabcolsep{2.4pt}
    \caption{Performance of \textsc{STILL-3-Tool-32B} and the baseline models on various mathematical tasks. Results with start (*) denote that they are copied from their reports or third-party website MathArena\protect\footnote{\url{https://matharena.ai/}}.}
      \begin{tabular}{lccccccc}
      \toprule
      \multirow{2.5}*{\textbf{Models}} & \multirow{2.5}*{\textbf{Tool}} & \multicolumn{2}{c}{\textbf{AIME 2024}} & \multicolumn{2}{c}{\textbf{AIME 2025}} & \multicolumn{2}{c}{\textbf{HMMT Feb. 2025}} \\
      \cmidrule(r){3-4}\cmidrule(r){5-6}\cmidrule(r){7-8}
      & & Greedy & Sample & Greedy & Sample & Greedy & Sample \\
      \midrule
      \textsc{STILL-2} & \ding{55} & 46.67 & - & - & - & - & - \\
      \textsc{QwQ-32B-Preview} & \ding{55} & - & 50.00* & - & 33.50* & - & - \\
      \textsc{DeepSeek R1} & \ding{55} & - & 79.80* & - & 70.00* & - & 42.00* \\
      \textsc{OpenAI O1} & \ding{55} & - & 79.20* & - & \textbf{79.00}* & - & 48.00* \\
      \textsc{OpenAI O3-mini-high} & \ding{55} & - & 79.60* & - & 76.50* & - & \textbf{53.00}* \\
      \midrule
      \textsc{DeepSeek-R1-Distill-Qwen-32B} & \ding{55} & 60.00 & 70.42 & 46.67 & 51.67 & 30.00 & 33.00 \\
      \textsc{Qwen2.5-Math-72B-Instruct}  & \ding{52} & 40.00 & - & - & - & - & - \\
      \midrule
      \midrule
      \multicolumn{8}{c}{\textsc{STILL-3-Tool-32B}} \\
      Synethsis (3k) & \ding{55} & 70.00 & 70.83 & 60.00 & 60.00 & 43.33 & 35.42  \\
      Distillation (0.8k) & \ding{55} & 73.33 & 73.33 & 56.67 & 58.75 & 40.00 & 43.75  \\
      Synethsis (1k) & \ding{52} & 76.67 & 74.58 & 50.00 & 60.42 & 43.33 & 45.83 \\
      Synethsis (3k) & \ding{52} & 80.00 & 72.08 & \textbf{66.67} & 65.00 & 43.33 & 47.50  \\
      Distillation (0.8k) & \ding{52} & \textbf{86.67} & \textbf{81.67} & 60.00 & 64.17 & \textbf{50.00} & 45.42  \\
      \bottomrule
      \end{tabular}
      \label{tab:still_3_tool_32B}
\end{table}

\subsection{\textsc{STILL-3-Tool-32B}: Empowering Reasoning Models with Tool Manipulation}
\label{sec-still-3-tool}

Tool manipulation represents a critical capability for enhancing and extending the functional scope of LLMs. Despite its potential, this ability has been largely underexplored in the context of reasoning models within existing literature. In this section, we aim to bridge this gap by equipping reasoning models with tool manipulation capabilities, thereby augmenting their problem-solving capacities and broadening their applicability.


\paratitle{Backbone Models.} To investigate this, we select \textsc{DeepSeek-R1-Distill-Qwen-32B} as the backbone model. The choice of a relatively large model size is motivated by the observation that tool manipulation can be challenging for smaller models, particularly when limited instruction data or computational resources are available. This aligns with findings from our prior work~\cite{still2}: smaller models struggle to learn slow-thinking reasoning patterns, whereas larger models can quickly acquire this ability even with minimal instruction data. Additionally, we focus on the code interpreter as the target tool, requiring the backbone model to generate accurate and executable programs—a task that poses significant challenges for smaller models.



\paratitle{Demonstration Data Distillation for SFT.} 
Before proceeding with RL experiments, we first conduct SFT experiments to systematically investigate the development of tool manipulation capabilities in a gradual and controlled manner. Specifically, we distill demonstration data that incorporates code snippets within the reasoning process.
{
We sample a subset with 3k questions from our whole question set (Section~\ref{sec-exp_setting}).
}
Once the question set is prepared, we employ powerful teacher models to generate responses that incorporate code execution. We primarily select \textsc{DeepSeek-R1} as the teacher model and adopt tool integration formats similar to those used in \textsc{ToRA}, \textsc{DeepSeekMath}, \textsc{Numina-TIR}, and \textsc{Qwen2.5-Math}. 
To distill training data from R1, we use a specific prompt to instruct R1 to solve problems with the assistance of code. After R1 generates a response, we verify whether it includes a code snippet and provides the correct answer. Only responses that meet both criteria are included in the training dataset. However, due to the high computational cost of distilling data directly from R1, we also synthesize training data from \textsc{DeepSeek-R1-Distill-Qwen-32B}, which offers a more cost-effective and efficient alternative. 
In our experiments, we initially attempted to use the same prompt to directly instruct \textsc{DeepSeek-R1-Distill-Qwen-32B} to generate solutions with code manipulation. However, we observed that the model struggled to follow these instructions effectively. We hypothesize that long CoT SFT process may have partially diminished its instruction-following capabilities. 
To address this, we employ a heuristic approach to refine the generated responses. Specifically, we first obtain rollouts from the model and split them into multiple parts. We then randomly select certain steps and inject a prefix (\eg ``\texttt{Wait a minute, maybe I could use some code to double-check my reasoning.}'') to simulate code integration, ensuring the responses align with our requirements.
Finally, we feed the prompt with code and results into \textsc{DeepSeek-R1-Distill-Qwen-32B} and guide it to complete the remaining thought and solution. This process can be iteratively to include more code snippets.

\paratitle{SFT Experiments.} 
We select \textsc{DeepSeek-R1-Distill-Qwen-32B} as the backbone model for fine-tuning, using a learning rate of $1 \times 10^{-5}$, a batch size of 96, and training for 17 epochs. To evaluate the model's performance, we employ three challenging benchmarks: AIME 2024, AIME 2025, and HMMT Feb. 2025. 
For a fair and comprehensive evaluation, we consider two sampling methods: greedy search (with a maximum token limit of 32768) and random sampling (with a temperature of $0.6$, top-$p$ sampling with $p=0.95$, a sample number of $n=8$, and a maximum token limit of $32768$). With these setups, we next present the experiment results and analysis. 

\begin{figure}[htbp]
    \centering
    \includegraphics[width=1.0\linewidth]{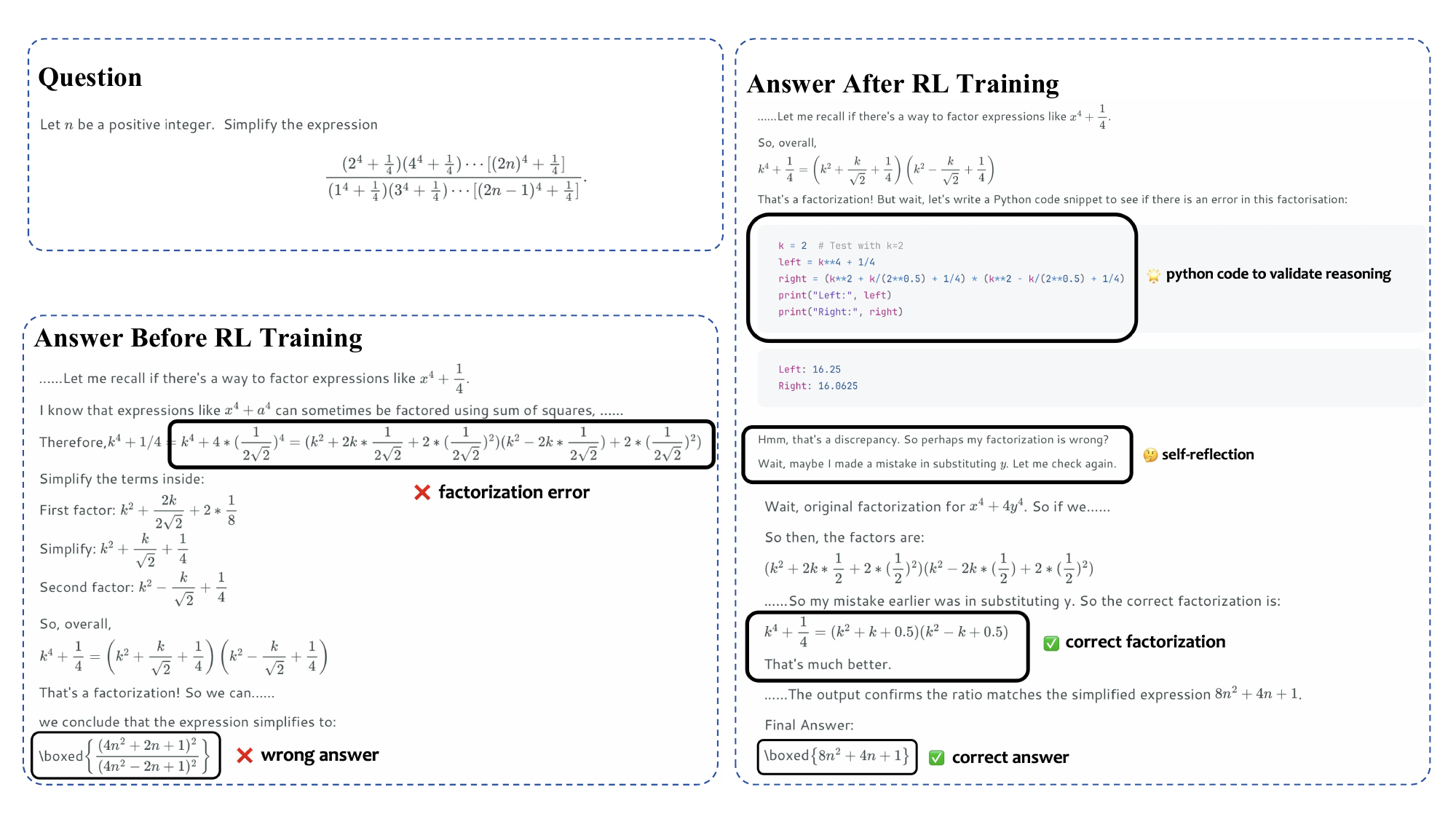} 
    \caption{A case study of \textsc{STILL-3-Tool-32B}.}
    \label{fig:still_3_tool_32B}
\end{figure}

$\bullet$ \emph{Performance Comparison}. 
In Table~\ref{tab:still_3_tool_32B}, we observe that \textsc{STILL-3-Tool-32B} outperforms all baseline models on AIME 2024, achieving a significant improvement of 15.56\%. This result underscores the effectiveness of integrating tool manipulation into the reasoning process. By training the model to think critically and autonomously invoke tools, our approach—developed based on \textsc{DeepSeek-R1-Distill-Qwen-32B}—achieves performance levels that are nearly comparable to DeepSeek-R1 across three major math competitions. 
These findings demonstrate that incorporating a code interpreter can substantially enhance the model's reasoning capabilities. Further analysis of the sampled responses (Figure~\ref{fig:still_3_tool_32B}) reveals that the model has effectively learned to invoke external tools when necessary, highlighting the effectiveness of our training methodology.

 
 
 

$\bullet$ \emph{Detailed Analysis}. 
We further analyze the fine-tuning results in detail. First, by comparing the teacher models—DeepSeek-R1 and the backbone model itself—we observe that the data distilled from R1 exhibits a higher quality. This is attributed to R1's superior capability in generating demonstration data that effectively incorporates tool use when prompted. Second, we find that a small amount of high-quality distilled data can suffice to elicit the backbone model's potential: the model variant trained on only 0.8k data instances distilled by DeepSeek-R1 achieves strong performance (a AIME score of 86.67 with greedy search).   
Third, we explore a variant where the model generates code snippets but does not invoke the code interpreter. While invoking the code interpreter significantly enhances model performance, we note that even without execution, generating code snippets still improves performance compared to the backbone model. This suggests that the process of generating code itself may encourage more structured and logical reasoning, even in the absence of tool execution.

These results demonstrate the effectiveness of generating code snippets and invoking external tools. For an illustrative example of \textsc{STILL-3-Tool-32B}, interested readers can refer to Figure~\ref{fig:still_3_tool_32B}.
A natural extension of this work is to leverage RL training to enhance the inherent tool manipulation capabilities of LLMs, a direction we leave for future research.



\section{Discussion}
\label{sec:discussion}

In this section, we address several key issues that are critical to the RL training process of slow-thinking LLMs and warrant further investigation.


\paratitle{Does Longer Response Refer to Higher Accuracy?}
Large reasoning models, capable of generating longer responses to facilitate intricate thinking and reasoning processes, have demonstrated remarkable performance across various complex reasoning tasks, such as mathematical problem-solving and coding challenges~\cite{deepseekr1,openaio1}. As previously discussed in this report (Section~\ref{fig:still_3_zero_32b}), longer responses  enable more deliberate reasoning, as the process essentially involves searching through a natural language space. However, prior studies~\cite{kimik1.5,Chen-arxiv-2024-Do} have identified a tendency in reasoning models to either ``overthink'' problems—such as redundantly checking irrelevant concepts or excessively verifying procedures—or ``underthink'' problems by frequently switching approaches without deliberate development.   
These behaviors do not lead to performance improvements and instead result in increased inference time and higher computational resource consumption.
Consistent with these findings, our experiments (Figure~\ref{fig:auxiliary_approach}) reveal that while response length continues to grow, accuracy remains stagnant in some settings. This suggests that the additional tokens in the responses do not translate into corresponding performance gains, indicating a decline in reasoning efficiency. The effectiveness of test-time scaling relies on the premise that increased token output leads to improved performance; if this relationship does not hold, the scaling effect becomes ineffective. 
To address this issue, a length penalty can be incorporated during the RL process to encourage models to compress the length of correct responses~\cite{kimik1.5,arora2025training}. While this approach enhances reasoning efficiency, it may also result in a trade-off with performance. A more ideal solution is to enable the model to adaptively allocate computational resources during test-time scaling, adjusting response length based on the difficulty of the query. Such an adaptive reasoning mode represents a promising direction for the development of reasoning models.


\paratitle{Does More Training Steps Lead to Better Performance?}
Although reasoning models primarily rely on test-time scaling to enhance performance, they are often developed through train-time compute scaling. This train-time scaling differs from the compute scaling used in pre-training; instead, it focuses on training processes designed to facilitate slow thinking, such as scaling RL training. The success of \textsc{DeepSeek-R1} demonstrates the positive impact of combining both train-time and test-time scaling. 
However, in our ``zero'' experiments, we observe a performance bottleneck as the number of training steps increases: after several hundred steps, the accuracy on both the training and test sets becomes increasingly difficult to improve. For instance, as shown in Figure~\ref{fig:para_comp}, the performance of \textsc{Qwen2.5-7B} on MATH-OAI and AIME 2024 datasets remains in a fluctuating state, struggling to achieve significant breakthroughs, such as surpassing 30\% accuracy on the AIME24 task—a threshold that can be readily attained through SFT on a distilled dataset.
The underlying cause of this phenomenon warrants further investigation. We hypothesize that a potential reason lies in the limited reasoning and exploration capabilities of the base model. Given current computational constraints, our RL training experiments have primarily focused on models with up to 32B parameters, which may encounter a capacity bottleneck that is challenging to overcome through RL training alone.   
Moreover, due to limited training budgets, we restricted the training to several hundred steps, which may be insufficient to surpass the performance bottleneck.  
Therefore, a question for future exploration is whether significantly extending the training steps, as in the case of \textsc{DeepSeek-R1}, would continue to constrain performance or instead lead to an emergent improvement in capabilities.

\begin{figure}[htbp]
    \centering
    \includegraphics[width=1.0\linewidth]{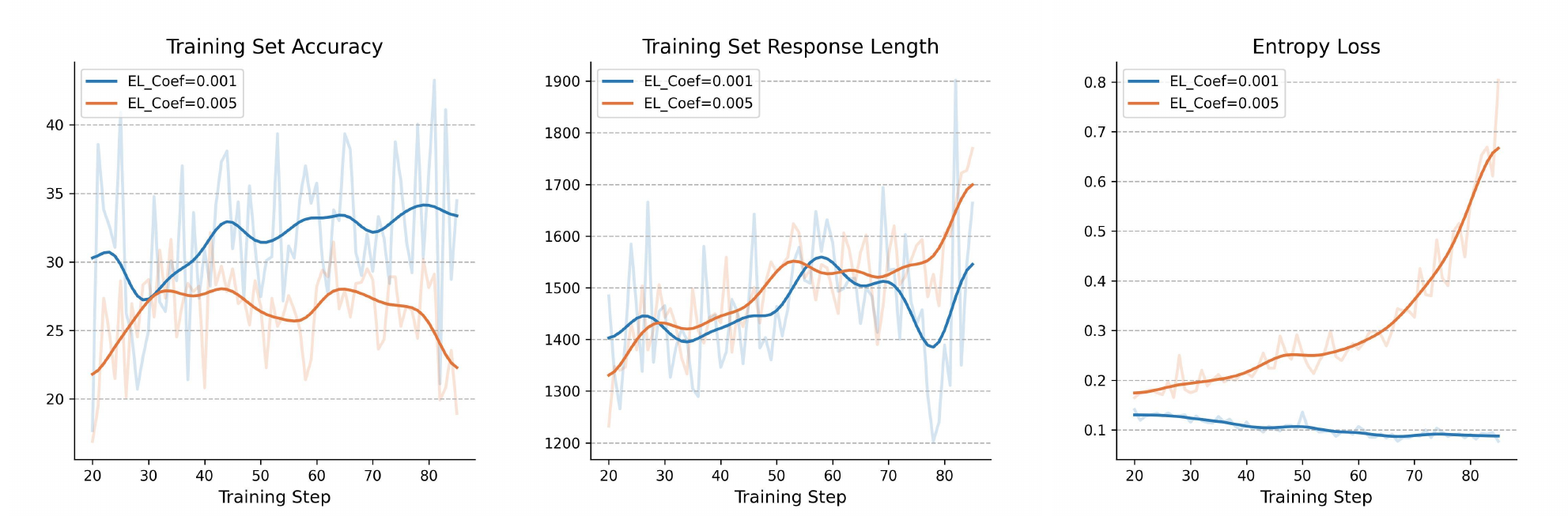} 
    \caption{The impact of different coefficients of entropy loss.}
    \label{fig:entropy_loss}
\end{figure}
\paratitle{Do LLMs Continue to {Develop Their Exploration Ability in the RL Training Process?}}
A key objective of RL training is to enhance the exploration capabilities of LLMs, enabling them to discover and expand the skill space of actions or strategies for complex reasoning. This represents a fundamental distinction between RL and SFT, which primarily relies on imitation learning. In theory, if a model progressively develops new capabilities through exploration, its performance on challenging problems should improve consistently. However, as discussed earlier (Figure~\ref{fig:para_comp}), we observe that model performance tends to stabilize or even decline as the number of training steps increases. This observation prompts us to critically examine the effectiveness of exploration during RL training. Specifically, we note that the diversity of the model's generated content diminishes rapidly and converges to a relatively low level.
Concretely, the entropy loss of LLMs decreases quickly in the initial stages of the RL process. A lower entropy loss corresponds to reduced diversity in the model's generated content. To mitigate this convergence, prior work has proposed incorporating an entropy bonus into the RL loss function~\cite{Hou-arxiv-2025-Advancing}. However, in our experiments (Figure~\ref{fig:entropy_loss}), we find that determining an appropriate coefficient for the entropy loss is empirically challenging.
{A coefficient that is too large can destabilize the model, while one that is too small only marginally slows the convergence of entropy loss without addressing the underlying issue.
This highlights an important issue for future research. We plan to investigate this issue further and encourage additional research to explore effective methods for balancing exploration and performance in RL training.
}


\paratitle{How Does RL Compare to SFT in Learning Paradigms?} Overall, SFT and RL represent two distinct learning paradigms for enhancing the capabilities of LLMs. 
In the SFT training process, the model rapidly improves its reasoning ability by imitating the behavior of a more powerful teacher model. In contrast, during RL training, the model refines its reasoning skills through continuous self-exploration and learning. By leveraging differences in reward scores between responses, the model learns to distinguish better from worse behaviors and optimizes its actions in parameter space to achieve higher rewards. 
Compared to SFT, RL has lower learning efficiency for improving model capacity but offers the advantage of autonomously expanding the scope of capabilities, not being limited to demonstrated behaviors. This paradigm is particularly well-suited for models with strong foundational capabilities, which are difficult to teach but more effectively guided through exploration. However, a key challenge in RL training is the difficulty of collecting sufficient high-quality reward signals to enhance model capacities. 
Existing reasoning models are primarily trained using verifiable rewards in a few limited domains, such as math, science, and coding. It would be intriguing to explore how this learning paradigm can be transferred to more general scenarios involving challenging tasks without definite answers, such as those targeted by Deep Research. 
\section{Conclusion}

In this paper, we presented a systematic study on reproducing R1-like reasoning models. Broadly, we categorized our experiments into two types: those based on base models (``zero'' experiments) and those based on fine-tuned models. For the first type of experiments, we extensively investigated various factors likely to influence RL performance, including training batch size, learning strategy, rollout parameters, and KL penalty. Building on these insights, we conducted ``zero'' training experiments on \textsc{Qwen2.5-32B}, which demonstrated a consistent increase in both model performance and response length.   
For the second type of experiments, we employed RL training to significantly enhance the performance of \textsc{DeepSeek-R1-Distill-Qwen-1.5B}, achieving an accuracy of 39 on AIME 2024. Additionally, we leveraged tool manipulation to further improve the reasoning capabilities of \textsc{DeepSeek-R1-Distill-Qwen-32B}, attaining an accuracy of 86.67 on AIME 2024 with greedy search.

Despite these algorithmic advancements, our work still lacks optimization in terms of efficiency.   
First, it is crucial to develop a train-inference hybrid framework to fully leverage computational resources. In RL training, sample rollout and parameter updates are alternated, and this process can be optimized for efficiency in various ways. Currently, we primarily rely on existing frameworks to achieve this, but we aim to explore further improvements in optimization efficiency. 
Second, long CoT reasoning poses significant challenges for developing efficient training and inference algorithms. For training, the acceleration of the entire process is primarily hindered by the lengthy rollouts generated by models. For inference, longer responses demand more computational resources per query, further complicating efficiency optimization.

As future work for our STILL project, we plan to delve deeper into the underlying mechanisms of deep reasoning and explore how RL algorithms can be trained more efficiently and stably on language models. Additionally, we aim to investigate the application of slow-thinking systems in real-world tasks, exploring the potential by enhancing the complex reasoning capabilities of LLMs.

\bibliographystyle{unsrt}
\bibliography{ref.bib}

\begin{thebibliography}{10}

\bibitem{zhao2023survey}
Wayne~Xin Zhao, Kun Zhou, Junyi Li, Tianyi Tang, Xiaolei Wang, Yupeng Hou, Yingqian Min, Beichen Zhang, Junjie Zhang, Zican Dong, Yifan Du, Chen Yang, Yushuo Chen, Zhipeng Chen, Jinhao Jiang, Ruiyang Ren, Yifan Li, Xinyu Tang, Zikang Liu, Peiyu Liu, Jian{-}Yun Nie, and Ji{-}Rong Wen.
\newblock A survey of large language models.
\newblock {\em CoRR}, abs/2303.18223, 2023.

\bibitem{deepseek-r1}
DeepSeek Team.
\newblock Deepseek-r1-lite-preview is now live: unleashing supercharged reasoning power!, November 2024.

\bibitem{kimik1.5}
Kimi Team, Angang Du, Bofei Gao, Bowei Xing, Changjiu Jiang, Cheng Chen, Cheng Li, Chenjun Xiao, Chenzhuang Du, Chonghua Liao, Chuning Tang, Congcong Wang, Dehao Zhang, Enming Yuan, Enzhe Lu, Fengxiang Tang, Flood Sung, Guangda Wei, Guokun Lai, Haiqing Guo, Han Zhu, Hao Ding, Hao Hu, Hao Yang, Hao Zhang, Haotian Yao, Haotian Zhao, Haoyu Lu, Haoze Li, Haozhen Yu, Hongcheng Gao, Huabin Zheng, Huan Yuan, Jia Chen, Jianhang Guo, Jianlin Su, Jianzhou Wang, Jie Zhao, Jin Zhang, Jingyuan Liu, Junjie Yan, Junyan Wu, Lidong Shi, Ling Ye, Longhui Yu, Mengnan Dong, Neo Zhang, Ningchen Ma, Qiwei Pan, Qucheng Gong, Shaowei Liu, Shengling Ma, Shupeng Wei, Sihan Cao, Siying Huang, Tao Jiang, Weihao Gao, Weimin Xiong, Weiran He, Weixiao Huang, Wenhao Wu, Wenyang He, Xianghui Wei, Xianqing Jia, Xingzhe Wu, Xinran Xu, Xinxing Zu, Xinyu Zhou, Xuehai Pan, Y.~Charles, Yang Li, Yangyang Hu, Yangyang Liu, Yanru Chen, Yejie Wang, Yibo Liu, Yidao Qin, Yifeng Liu, Ying Yang, Yiping Bao, Yulun Du, Yuxin Wu, Yuzhi Wang, Zaida Zhou,
  Zhaoji Wang, Zhaowei Li, Zhen Zhu, Zheng Zhang, Zhexu Wang, Zhilin Yang, Zhiqi Huang, Zihao Huang, Ziyao Xu, and Zonghan Yang.
\newblock Kimi k1.5: Scaling reinforcement learning with llms.
\newblock {\em CoRR}, abs/2501.12599, 2025.

\bibitem{jiang2024technical}
Jinhao Jiang, Zhipeng Chen, Yingqian Min, Jie Chen, Xiaoxue Cheng, Jiapeng Wang, Yiru Tang, Haoxiang Sun, Jia Deng, Wayne~Xin Zhao, et~al.
\newblock Technical report: Enhancing llm reasoning with reward-guided tree search.
\newblock {\em CoRR}, abs/2411.11694, 2024.

\bibitem{qiu-o1-survey}
Zhiyuan Zeng, Qinyuan Cheng, Zhangyue Yin, Bo~Wang, Shimin Li, Yunhua Zhou, Qipeng Guo, Xuanjing Huang, and Xipeng Qiu.
\newblock Scaling of search and learning: {A} roadmap to reproduce o1 from reinforcement learning perspective.
\newblock {\em CoRR}, abs/2412.14135, 2024.

\bibitem{learn-to-summarize}
Nisan Stiennon, Long Ouyang, Jeff Wu, Daniel~M. Ziegler, Ryan Lowe, Chelsea Voss, Alec Radford, Dario Amodei, and Paul~F. Christiano.
\newblock Learning to summarize from human feedback.
\newblock {\em CoRR}, abs/2009.01325, 2020.

\bibitem{Ouyang2022instruct}
Long Ouyang, Jeffrey Wu, Xu~Jiang, Diogo Almeida, Carroll~L. Wainwright, Pamela Mishkin, Chong Zhang, Sandhini Agarwal, Katarina Slama, Alex Ray, John Schulman, Jacob Hilton, Fraser Kelton, Luke Miller, Maddie Simens, Amanda Askell, Peter Welinder, Paul~F. Christiano, Jan Leike, and Ryan Lowe.
\newblock Training language models to follow instructions with human feedback.
\newblock In {\em NeurIPS}, 2022.

\bibitem{still2}
Yingqian Min, Zhipeng Chen, Jinhao Jiang, Jie Chen, Jia Deng, Yiwen Hu, Yiru Tang, Jiapeng Wang, Xiaoxue Cheng, Huatong Song, et~al.
\newblock Imitate, explore, and self-improve: A reproduction report on slow-thinking reasoning systems.
\newblock {\em arXiv preprint arXiv:2412.09413}, 2024.

\bibitem{deepseekr1}
DeepSeek{-}AI, Daya Guo, Dejian Yang, Haowei Zhang, Junxiao Song, Ruoyu Zhang, Runxin Xu, Qihao Zhu, Shirong Ma, Peiyi Wang, Xiao Bi, Xiaokang Zhang, Xingkai Yu, Yu~Wu, Z.~F. Wu, Zhibin Gou, Zhihong Shao, Zhuoshu Li, Ziyi Gao, Aixin Liu, Bing Xue, Bingxuan Wang, Bochao Wu, Bei Feng, Chengda Lu, Chenggang Zhao, Chengqi Deng, Chenyu Zhang, Chong Ruan, Damai Dai, Deli Chen, Dongjie Ji, Erhang Li, Fangyun Lin, Fucong Dai, Fuli Luo, Guangbo Hao, Guanting Chen, Guowei Li, H.~Zhang, Han Bao, Hanwei Xu, Haocheng Wang, Honghui Ding, Huajian Xin, Huazuo Gao, Hui Qu, Hui Li, Jianzhong Guo, Jiashi Li, Jiawei Wang, Jingchang Chen, Jingyang Yuan, Junjie Qiu, Junlong Li, J.~L. Cai, Jiaqi Ni, Jian Liang, Jin Chen, Kai Dong, Kai Hu, Kaige Gao, Kang Guan, Kexin Huang, Kuai Yu, Lean Wang, Lecong Zhang, Liang Zhao, Litong Wang, Liyue Zhang, Lei Xu, Leyi Xia, Mingchuan Zhang, Minghua Zhang, Minghui Tang, Meng Li, Miaojun Wang, Mingming Li, Ning Tian, Panpan Huang, Peng Zhang, Qiancheng Wang, Qinyu Chen, Qiushi Du, Ruiqi Ge,
  Ruisong Zhang, Ruizhe Pan, Runji Wang, R.~J. Chen, R.~L. Jin, Ruyi Chen, Shanghao Lu, Shangyan Zhou, Shanhuang Chen, Shengfeng Ye, Shiyu Wang, Shuiping Yu, Shunfeng Zhou, Shuting Pan, and S.~S. Li.
\newblock Deepseek-r1: Incentivizing reasoning capability in llms via reinforcement learning.
\newblock {\em CoRR}, abs/2501.12948, 2025.

\bibitem{openrlhf}
Jian Hu, Xibin Wu, Zilin Zhu, Xianyu, Weixun Wang, Dehao Zhang, and Yu~Cao.
\newblock Openrlhf: An easy-to-use, scalable and high-performance rlhf framework.
\newblock {\em arXiv preprint arXiv:2405.11143}, 2024.

\bibitem{verl}
Guangming Sheng, Chi Zhang, Zilingfeng Ye, Xibin Wu, Wang Zhang, Ru~Zhang, Yanghua Peng, Haibin Lin, and Chuan Wu.
\newblock Hybridflow: A flexible and efficient rlhf framework.
\newblock {\em arXiv preprint arXiv: 2409.19256}, 2024.

\bibitem{qwen2.5}
An~Yang, Baosong Yang, Beichen Zhang, Binyuan Hui, Bo~Zheng, Bowen Yu, Chengyuan Li, Dayiheng Liu, Fei Huang, Haoran Wei, Huan Lin, Jian Yang, Jianhong Tu, Jianwei Zhang, Jianxin Yang, Jiaxi Yang, Jingren Zhou, Junyang Lin, Kai Dang, Keming Lu, Keqin Bao, Kexin Yang, Le~Yu, Mei Li, Mingfeng Xue, Pei Zhang, Qin Zhu, Rui Men, Runji Lin, Tianhao Li, Tingyu Xia, Xingzhang Ren, Xuancheng Ren, Yang Fan, Yang Su, Yichang Zhang, Yu~Wan, Yuqiong Liu, Zeyu Cui, Zhenru Zhang, and Zihan Qiu.
\newblock Qwen2.5 technical report.
\newblock {\em CoRR}, abs/2412.15115, 2024.

\bibitem{dan2021math}
Dan Hendrycks, Collin Burns, Saurav Kadavath, Akul Arora, Steven Basart, Eric Tang, Dawn Song, and Jacob Steinhardt.
\newblock Measuring mathematical problem solving with the {MATH} dataset.
\newblock In Joaquin Vanschoren and Sai{-}Kit Yeung, editors, {\em Proceedings of the Neural Information Processing Systems Track on Datasets and Benchmarks 1, NeurIPS Datasets and Benchmarks 2021, December 2021, virtual}, 2021.

\bibitem{li2024numinamath}
Jia Li, Edward Beeching, Lewis Tunstall, Ben Lipkin, Roman Soletskyi, Shengyi Huang, Kashif Rasul, Longhui Yu, Albert~Q Jiang, Ziju Shen, et~al.
\newblock Numinamath: The largest public dataset in ai4maths with 860k pairs of competition math problems and solutions.
\newblock {\em Hugging Face repository}, 2024.

\bibitem{orz}
Jingcheng Hu, Yinmin Zhang, Qi~Han, Daxin Jiang, and Heung-Yeung~Shum Xiangyu~Zhang.
\newblock Open-reasoner-zero: An open source approach to scaling reinforcement learning on the base model.
\newblock \url{https://github.com/Open-Reasoner-Zero/Open-Reasoner-Zero}, 2025.

\bibitem{aaron2017sympy}
Aaron Meurer, Christopher~P. Smith, Mateusz Paprocki, Ondrej Cert{\'{\i}}k, Sergey~B. Kirpichev, Matthew Rocklin, Amit Kumar, Sergiu Ivanov, Jason~Keith Moore, Sartaj Singh, Thilina Rathnayake, Sean Vig, Brian~E. Granger, Richard~P. Muller, Francesco Bonazzi, Harsh Gupta, Shivam Vats, Fredrik Johansson, Fabian Pedregosa, Matthew~J. Curry, Andy~R. Terrel, Step{\'{a}}n Roucka, Ashutosh Saboo, Isuru Fernando, Sumith Kulal, Robert Cimrman, and Anthony~M. Scopatz.
\newblock Sympy: symbolic computing in python.
\newblock {\em PeerJ Comput. Sci.}, 3:e103, 2017.

\bibitem{OmniMATH}
Bofei Gao, Feifan Song, Zhe Yang, Zefan Cai, Yibo Miao, Qingxiu Dong, Lei Li, Chenghao Ma, Liang Chen, Runxin Xu, Zhengyang Tang, Benyou Wang, Daoguang Zan, Shanghaoran Quan, Ge~Zhang, Lei Sha, Yichang Zhang, Xuancheng Ren, Tianyu Liu, and Baobao Chang.
\newblock Omni-math: {A} universal olympiad level mathematic benchmark for large language models.
\newblock {\em CoRR}, abs/2410.07985, 2024.

\bibitem{aopsdataset}
Sadegh Mahdavi, Muchen Li, Kaiwen Liu, Christos Thrampoulidis, Leonid Sigal, and Renjie Liao.
\newblock Leveraging online olympiad-level math problems for llms training and contamination-resistant evaluation, 2025.

\bibitem{matharena}
Mislav Balunović, Jasper Dekoninck, Ivo Petrov, Nikola Jovanović, and Martin Vechev.
\newblock Matharena: Evaluating llms on uncontaminated math competitions, February 2025.

\bibitem{Hou-arxiv-2025-Advancing}
Zhenyu Hou, Xin Lv, Rui Lu, Jiajie Zhang, Yujiang Li, Zijun Yao, Juanzi Li, Jie Tang, and Yuxiao Dong.
\newblock Advancing language model reasoning through reinforcement learning and inference scaling.
\newblock {\em CoRR}, abs/2501.11651, 2025.

\bibitem{openthoughts}
OpenThoughts Team.
\newblock {Open Thoughts}.
\newblock https://open-thoughts.ai, January 2025.

\bibitem{openaio1}
Aaron Jaech, Adam Kalai, Adam Lerer, Adam Richardson, Ahmed El{-}Kishky, Aiden Low, Alec Helyar, Aleksander Madry, Alex Beutel, Alex Carney, Alex Iftimie, Alex Karpenko, Alex~Tachard Passos, Alexander Neitz, Alexander Prokofiev, Alexander Wei, Allison Tam, Ally Bennett, Ananya Kumar, Andre Saraiva, Andrea Vallone, Andrew Duberstein, Andrew Kondrich, Andrey Mishchenko, Andy Applebaum, Angela Jiang, Ashvin Nair, Barret Zoph, Behrooz Ghorbani, Ben Rossen, Benjamin Sokolowsky, Boaz Barak, Bob McGrew, Borys Minaiev, Botao Hao, Bowen Baker, Brandon Houghton, Brandon McKinzie, Brydon Eastman, Camillo Lugaresi, Cary Bassin, Cary Hudson, Chak~Ming Li, Charles de~Bourcy, Chelsea Voss, Chen Shen, Chong Zhang, Chris Koch, Chris Orsinger, Christopher Hesse, Claudia Fischer, Clive Chan, Dan Roberts, Daniel Kappler, Daniel Levy, Daniel Selsam, David Dohan, David Farhi, David Mely, David Robinson, Dimitris Tsipras, Doug Li, Dragos Oprica, Eben Freeman, Eddie Zhang, Edmund Wong, Elizabeth Proehl, Enoch Cheung, Eric Mitchell,
  Eric Wallace, Erik Ritter, Evan Mays, Fan Wang, Felipe~Petroski Such, Filippo Raso, Florencia Leoni, Foivos Tsimpourlas, Francis Song, Fred von Lohmann, Freddie Sulit, Geoff Salmon, Giambattista Parascandolo, Gildas Chabot, Grace Zhao, Greg Brockman, Guillaume Leclerc, Hadi Salman, Haiming Bao, Hao Sheng, Hart Andrin, Hessam Bagherinezhad, Hongyu Ren, Hunter Lightman, Hyung~Won Chung, Ian Kivlichan, Ian O'Connell, Ian Osband, Ignasi~Clavera Gilaberte, and Ilge Akkaya.
\newblock Openai o1 system card.
\newblock {\em CoRR}, abs/2412.16720, 2024.

\bibitem{Chen-arxiv-2024-Do}
Xingyu Chen, Jiahao Xu, Tian Liang, Zhiwei He, Jianhui Pang, Dian Yu, Linfeng Song, Qiuzhi Liu, Mengfei Zhou, Zhuosheng Zhang, Rui Wang, Zhaopeng Tu, Haitao Mi, and Dong Yu.
\newblock Do {NOT} think that much for 2+3=? on the overthinking of o1-like llms.
\newblock {\em CoRR}, abs/2412.21187, 2024.

\bibitem{arora2025training}
Daman Arora and Andrea Zanette.
\newblock Training language models to reason efficiently.
\newblock {\em arXiv preprint arXiv:2502.04463}, 2025.

\end{thebibliography}

\end{document}